\documentclass[11pt]{article}

\usepackage[preprint]{acl}

\usepackage{times}
\usepackage{latexsym}
\usepackage[disable]{todonotes} 
\usepackage{tcolorbox}
\tcbuselibrary{breakable}
\usepackage[T1]{fontenc}

\usepackage[utf8]{inputenc}

\usepackage{microtype}

\usepackage{inconsolata}

\usepackage{graphicx}

\usepackage{booktabs}
\usepackage{amsmath}
\usepackage{amssymb}
\usepackage{subcaption}
\usepackage{tabularx}

\usepackage{colortbl}  
\definecolor{rowgray}{gray}{0.92}
\usepackage{enumitem}

\title{What Gets Unmasked First? Trajectory Analysis of Diffusion Models for Graph-to-Text Generation}

\author{Qing Wang\thanks{Equal contribution authors}, Jacob Devasier\footnotemark[1], Chengkai Li \\
        The University of Texas at Arlington \\
        \{qxw5305, jacob.devasier\}@mavs.uta.edu, cli@uta.edu}

\begin{document}
\maketitle
\begin{abstract}
We present the first systematic study of masked diffusion language models (MDLMs) for graph-to-text generation. We analyze MDLM generation trajectories---the order in which tokens are unmasked during iterative decoding---and find that, unlike autoregressive LLMs which generate text linearly, MDLMs naturally prioritize entities first, followed by relational and function words, with structural tokens resolved last. We further identify a previously undocumented failure mode of supervised fine-tuning: SFT disrupts this strategy by prematurely anchoring structural sentence-ending tokens early in the decoding trajectory, effectively fixing the output length which can lead to omitted or hallucinated information. To address this, we propose $\lambda$-scaled structural decoding, a training-free inference-time modification that downweights structural token confidence and recovers +9.4 BLEU-4. Finally, we introduce Graph-LLaDA, which integrates a Graph Transformer encoder into LLaDA's decoding process to explicitly incorporate relational graph structure. Cross-dataset evaluation on LAGRANGE reveals that previous baselines overfit to dataset-specific patterns, while LLM- and MDLM-based approaches generalize significantly better.

\end{abstract}

\section{Introduction}


Knowledge graphs encode information as structured networks of entities and their relations. Generating natural language descriptions from such graphs requires a model to interpret this relational structure and render it as fluent, faithful text. Autoregressive large language models (LLMs) have demonstrated strong performance across many natural language generation tasks and have become the dominant approach for graph-to-text generation, achieving strong results on benchmarks such as WebNLG~\cite{castro-ferreira20:bilin-bi-direc-webnl-shared,ribeiro-etal-2021-investigating,he2025plangtg}. However, text-only autoregressive models struggle to fully capture the structure of graphs, often leading to information omission or hallucination~\cite{shi-etal-2023-hallucination}. 

These models generate text one token at a time in a fixed left-to-right order, meaning they must commit to a linearization of the graph structure from the very first token, without the ability to revisit or revise earlier choices. Furthermore, prior work has shown that autoregressive LLMs frequently struggle with complex graph planning, triplet ordering, and factual grounding, especially when the number of graph triplets increases~\cite{yuan-faerber-2023-evaluating,he-etal-2025-evaluating-improving}. Whether this sequential constraint is merely an inconvenience or a fundamental limitation for graph-to-text generation remains an open question.

Masked diffusion language models (MDLMs) \cite{austin2021d3pm,sahoo2024mdlm} offer a fundamentally different generation process. Rather than producing tokens sequentially, MDLMs such as LLaDA~\cite{nie2026large} and Dream-7B~\cite{ye2025dream} begin with a fully masked sequence and iteratively unmask tokens over multiple refinement steps. Unlike autoregressive models, MDLMs select which positions to resolve at each step, making irreversible commitments in a flexible, content-dependent order rather than a predetermined sequential one. This flexibility is intuitively well aligned with how humans approach the same task: when translating a graph into text, we typically first identifies which entities are present and how they relate before composing full sentences. In principle, an MDLM could mirror this process, first establishing which entities to mention and then filling in the relational language connecting them. 

Despite this potential, MDLMs remain largely unexplored in the graph-to-text generation literature. It is unclear how they construct text from graph-structured inputs, whether their generation trajectories exhibit systematic patterns, and whether they offer practical advantages over autoregressive LLMs on this task. To address these questions, we apply LLaDA to the WebNLG~\cite{web_nlg} and LAGRANGE~\cite{mousavi-etal-2024-construction} benchmarks and conduct a detailed analysis of its generation trajectory---the order in which tokens are predicted over the course of iterative unmasking. Our analysis reveals that LLaDA follows a clear content-first generation pattern: it first predicts entity tokens, then resolves relational and function-word tokens together, and finally commits structural tokens such as punctuation. The entity-first behavior replicates across both benchmarks and on another MDLM, Dream-7B~\cite{ye2025dream}, indicating that it is a property of masked diffusion rather than of a particular backbone or benchmark. 

Our trajectory analysis also uncovers a previously unidentified failure mode: after supervised fine-tuning (SFT), LLaDA tends to commit structural tokens, such as the period token (``.'') and the end-of-sentence (EOS) token, very early in the denoising process, before substantive content has been resolved. To mitigate premature structural commitment, we propose \emph{$\lambda$-scaled structural decoding}, a training-free inference-time modification that downweights the confidence of structural tokens, delaying their commitment and substantially improving generation quality.

Motivated by these findings, we propose \textbf{Graph-LLaDA}, which augments the LLaDA generation pipeline with a Graph Transformer~\cite{yun2019graph} encoder that explicitly incorporates graph structure into the diffusion process. The encoder produces structure-aware node representations that are injected into LLaDA's input sequence alongside the linearized triples, enabling the model to attend to both the relational topology and the lexical content of the input graph.

Our main contributions are as follows:
\begin{itemize}[noitemsep,wide,topsep=0pt]
    \item We present the first systematic study of masked diffusion language models (MDLMs) for graph-to-text generation and establish strong baselines on WebNLG and LAGRANGE. 
    
    \item We conduct a detailed analysis of MDLM generation trajectories on graph-to-text, characterizing the order in which the denoising process commits to entities, relations, and structural tokens.
    
    \item We identify premature structural commitment as a failure mode of SFT-trained MDLMs and propose $\lambda$-scaled structural decoding to mitigate it.
    
    \item We propose \textbf{Graph-LLaDA}, which augments the LLaDA generation pipeline with a Graph Transformer encoder that explicitly incorporates graph structure into the decoding process.
\end{itemize}
\section{Methodology}
\label{sec:methodology}

We first define the graph-to-text generation task, then present our methodology for analyzing the generation trajectories of masked diffusion models, propose a decoding modification motivated by this analysis, and finally describe how we extend LLaDA with a graph encoder to incorporate structural information from knowledge graphs.





\subsection{Problem Formulation}
Graph-to-text generation is the task of automatically generating fluent and faithful natural language text from structured graph inputs. Formally, each triple $t = (s_i, r_i, o_i)$ represents a directed edge in the graph, where $s_i$, $r_i$, $o_i$ denote the subject, relation, and object, respectively. A set of such triples $S = \{t_1,t_2,\ldots,t_n\}$ forms a directed graph that encodes structured factual knowledge. Given $S$ as input, the goal is to generate a natural language description $D$ that accurately and completely conveys the semantic information encoded in the graph.

\begin{figure}[t]
    \centering
   \includegraphics[width=\linewidth]{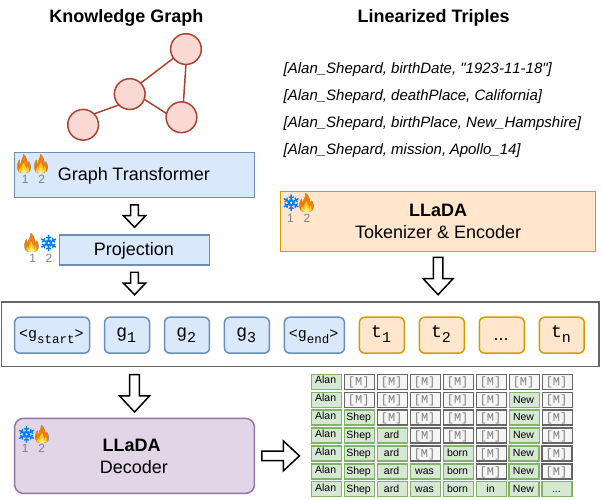}
  \caption{An overview of Graph-LLaDA. Trained and frozen modules of our two-stage training process are indicated along with their associated stage. }
  \label{fig:graph-llada}
\end{figure}

\subsection{Generation Trajectory Analysis} 
LLaDA is a masked diffusion language model that generates text by iteratively unmasking a fully masked sequence, treating mask tokens as the noise to be removed. At each denoising step, the model predicts all masked positions simultaneously using bidirectional attention and commits the most confident predictions. This process repeats until every token is unmasked. A central question of this work is whether masked diffusion models follow systematic patterns when constructing text from graph-structured inputs---specifically, whether core elements such as entities and relations are resolved at particular stages of the diffusion trajectory. To answer this question, we record the complete unmasking trajectory of LLaDA and Dream-7B during inference. 

 
\paragraph{Trajectory Recording.}
We record each committed token along with its position in the output sequence and the denoising timestep at which it was committed. We normalize each token's commitment timestep to $[0, 1]$, where $0$ corresponds to the first denoising step and $1$ to the final step. This allows trajectory statistics to be compared across generations of different lengths and across models that use different numbers of denoising steps. The resulting trajectory provides a complete temporal record of the generation process, capturing when each output token was decided.

\paragraph{Token Classification.}

To analyze whether the generation trajectory reflects the structure of the input graph, each generated token is classified into one of five categories: \textit{entity}, \textit{relation}, \textit{stop word}, \textit{punctuation}, or \textit{other}. Entity and relation tokens correspond to content derived from the input triples, stop words capture function words, punctuation captures sentence-level delimiters, and the \textit{other} category contains tokens --- typically connective phrasing or paraphrastic content --- that could not be mapped into any of the preceding categories. In practice, \textit{other} tokens constitute a small minority of all generated tokens and are heavily concentrated at the end of the trajectory, so they do not undermine our analysis (see Table~\ref{tab:trajectory}). We apply a priority-based procedure in which each token is tested against a sequence of classification criteria, and the first match determines its label. Details of this procedure can be found in Appendix~\ref{app:token-classification}.

\subsection{$\lambda$-Scaled Structural Decoding}
\label{sec:lambda-scaling}
A second observation from our trajectory analysis motivates an inference-time intervention. After supervised fine-tuning, LLaDA shifts toward committing structural tokens---sentence-terminating punctuation, end-of-sequence, and end-of-turn markers---very early in the denoising process, before the bulk of the content has been resolved. This premature commitment effectively fixes the length and segmentation of the output in advance, producing generations that are either padded with redundant content or truncated. We defer the supporting quantitative evidence---first-period timing, content-before-period proportions, and per-$\lambda$ recovery curves---to Section~\ref{sec:exp-traj-analysis}. To the best of our knowledge, this failure mode has not been previously identified, and we view its diagnosis as one of the main contributions of this work: it isolates a concrete mechanism by which SFT can degrade generation quality in MDLMs despite improving token-level accuracy.



To address this premature structural commitment, we introduce a simple inference-time modification that downweights the confidence of structural tokens. At each denoising step, before tokens are committed, we multiply the predicted confidence of any structural token by a scalar $\lambda \in (0, 1]$. This causes structural tokens to require higher raw confidence before being selected (or rather it requires that all other tokens have sufficiently low confidence), delaying their commitment relative to content tokens and reducing the risk of premature sequence termination. A potential alternative is using an MDLM supporting flexible-length generation~\cite{kim2025any}; however, their approach requires significant additional training, while our modification requires no retraining and can be applied to any MDLM at inference time.

\subsection{Graph-Augmented LLaDA}
\label{sec:graph-llada}

Orthogonal to the decoding modification above, we also investigate whether explicitly encoding graph structure improves generation quality. We propose Graph-LLaDA, which extends LLaDA with a Graph Neural Network (GNN) encoder that provides structural information during generation. We chose LLaDA as the base model because it was the first masked diffusion language model scaled to competitive performance with many autoregressive models. Dream-7B~\cite{ye2025dream} was another candidate for this work; however, as shown in Table~\ref{tab:webnlg}, its zero-shot performance was significantly lower than LLaDA. LLaDA 2~\cite{bie2025llada2} was also considered but was not chosen due to its significantly larger compute requirements. Our approach does not place any constraints on model architecture or selection and can in principle be extended to other MDLM backbones.

As illustrated in Figure~\ref{fig:graph-llada}, the model receives two parallel representations of the input knowledge graph: a \textbf{graph stream} that encodes relational topology through learned embeddings, and a \textbf{text stream} that includes the raw triples as linearized text for lexical grounding. This dual representation allows the GNN to capture graph structure while ensuring that entities and relations remain directly accessible to the language model.

\paragraph{Graph Encoding.}
The input triples are parsed into a directed graph where entities are nodes and relations are edges. Node and edge features are initialized by tokenizing each entity or relation with LLaDA's tokenizer, retrieving the corresponding pretrained token embeddings, and mean-pooling over the token sequence. This produces a semantically meaningful initialization grounded in LLaDA's representations. Since LLaDA's embedding dimension is larger than what the GNN requires, a learned linear projection maps the pooled embeddings down to the GNN's hidden size $d_g$. Edge features are derived the same way through an analogous projection.

The resulting graph is then encoded using a Graph Transformer, which updates node representations by attending over their neighborhoods. Crucially, edge features participate in this attention mechanism: relation embeddings are used to modulate the attention weights between connected nodes, so that the model can distinguish, for example, a \textit{birthPlace} edge from a \textit{nationality} edge between the same pair of entities. This is particularly important for graph-to-text generation, where relations often correspond to the main verbs or predicative content of the target sentence, and faithfully preserving their semantics is essential for factual accuracy. Finally, a learned MLP projects the GNN's output embeddings back into LLaDA's embedding space, so that the structure-aware node representations are semantically aligned and dimensionally consistent with LLaDA's token embeddings and can be directly inserted into the sequence.
\todo{MLP and the node and edge projections are not in the diagram. if you have time, it could make sense to add such details.}

\paragraph{Embedding Injection.}
The text prompt includes a \texttt{<graph>} marker that expands into a sequence of $N$ placeholder tokens (one per node in the input graph), delimited by \texttt{<g\_start>} and \texttt{<g\_end>} markers. After tokenization, each placeholder position is replaced with the corresponding projected node embedding from the GNN.

We adopt this per-node injection strategy, following GraphGPT~\citep{tang2024graphgpt}, rather than compressing the entire graph into a single token as in G-Retriever~\citep{he2024gretriever}. Per-node tokens allow the model to attend to individual entity representations, which is particularly important for multi-hop reasoning over knowledge graphs. Recent work~\citep{grover2026one} has confirmed that single-token pooling creates an information bottleneck and that increasing the number of graph tokens recovers the lost performance.

The prompt additionally contains the raw triples rendered as bracketed tuples (e.g., \texttt{[subject, relation, object]}), providing lexical grounding alongside the structural signal. During both training and inference, the graph token positions are never masked by LLaDA's forward diffusion process and are excluded from the diffusion loss.


\paragraph{Training.} 

We follow a two-stage training strategy similar to the approach of~\citet{tang2024graphgpt}. In stage~\textbf{(1)}, we freeze the LLM and train only the GNN and the projection layers. This forces the projector to learn a mapping from the GNN's output space into LLaDA's embedding space, aligning the two representations. In stage~\textbf{(2)}, we freeze the GNN and train the projection layer alongside LoRA~\cite{hu2022lora} weights on the LLM, allowing the language model to adapt to the graph-conditioned inputs.

\section{Experimental Setup}

\paragraph{Datasets.}

We evaluate on two benchmarks. Our primary benchmark is WebNLG v3.0~\cite{castro-ferreira20:bilin-bi-direc-webnl-shared}, a manually curated dataset for graph-to-text generation. Each input consists of 1--7 RDF triples extracted from DBpedia, paired with one or more reference descriptions. We train on the WebNLG training split and evaluate on the English test set, which contains 1,779 input triple sets. All results are reported using multi-reference evaluation. Although WebNLG is a high-quality benchmark, its ontology coverage is limited and its graph structures are relatively constrained~\cite{shi-etal-2023-hallucination}.

To assess generalization beyond WebNLG's domain, we additionally evaluate on LAGRANGE~\cite{mousavi-etal-2024-construction}, a large-scale graph-text aligned corpus created by aligning Wikipedia sentences with Wikidata. Models are not trained on LAGRANGE; we evaluate using either zero-shot inference or models already trained on WebNLG. The one exception is GAP, which we also train on LAGRANGE to examine whether its strong in-domain performance transfers out of domain on WebNLG.\todo[color=orange]{GAP coming out of nowhere.}

\paragraph{Models.}

We adopt LLaDA-8B as our primary model to investigate how masked diffusion language models handle entities and relational structures in graph-to-text generation. For MDLMs, we also evaluate Dream-7B~\cite{ye2025dream} in a zero-shot setting.\todo{need to smoothen thw description. not sure why it says "For MDLMs". Isn't LLaDA itself already an MDLM?} For autoregressive baselines, we evaluate Qwen3-8B~\cite{qwen3technicalreport}, LLaMA~3.1-8B~\cite{grattafiori2024llama}, and LLaMA~3.3-70B under both zero-shot and LoRA-based supervised fine-tuning. For supervised baselines, we compare against GAP~\cite{colas-etal-2022-gap}, JointGT~\cite{ke-etal-2021-jointgt}, and MGSA~\cite{wang2024mgsa}. We retrain GAP ourselves to ensure a fair comparison; results for the remaining supervised baselines are taken from published numbers.\todo{why do we only retrain GAP?}\todo{compare what "against GAP, ..."?}

\paragraph{Implementation.}
All models are fine-tuned with LoRA; full hyperparameters and training details are provided in Appendix~\ref{app:multi-token-generation}. Graph-LLaDA uses a Graph Transformer encoder followed by a 2-layer MLP projector, trained in two stages as described in Section~\ref{sec:graph-llada}. Unless otherwise noted, we apply $\lambda$-scaled structural decoding with $\lambda=0.25$ at inference time. This is chosen as it offered the best performance, as shown in Table~\ref{tab:alpha}.\todo{not ideal to mwntion the table this way. this comment is more for future revision.} We use a shared prompt template across all models without per-model tuning (Appendix~\ref{app:prompts}).

\todo{probably unnecessary to make evaluation its own subsection. it is strange to just have one subsection.}

\subsection{Evaluation.}

\paragraph{Metrics.}
We report four automatic metrics: BLEU~\cite{bleu}, METEOR~\cite{meteor}, ROUGE-L~\cite{lin-2004-rouge}, and CIDEr~\cite{wang2024mgsa}. These capture lexical overlap\todo{overlap between what and what?} with the reference descriptions but cannot fully reflect semantic quality or detect hallucinations, so we complement them with two preference-based evaluations. 


\paragraph{Human Evaluation.}
First, we conduct a human evaluation on a random sample of 50 WebNLG outputs from our best model. Two English speakers independently rate each output on three dimensions: fluency, hallucinations, and omissions. The samples are uniformly distributed by KG size\todo{i suppose you meant 'graph size'. The graph in each test case is not a KG itself.} to allow comparison across input complexity. We additionally ask Claude Sonnet~4.6 and GPT~5.5 to perform the same annotation\todo{strange to call it annotation.} for comparison with the human judgments.\todo{is this the same as LlM-judge evaluation?} Results are reported in Table~\ref{tab:human-analysis}.\todo{i thought we have space. why not includ and explain the table here?}

\paragraph{LLM-Judge Evaluation.}
Second, we scale the preference evaluation across all analyzed models using an LLM judge in a pairwise head-to-head setup. For each model pair, the judge evaluates 1,000 outputs drawn from both WebNLG and LAGRANGE, rating each comparison on the same three dimensions. We use two judges: Gemma~4 31B~\cite{gemma4_2026} and Qwen~3.6~\cite{qwen36_35b_a3b}. To mitigate positional bias, we evaluate each pair twice with the output order reversed, with preference retained only when the judge's decision is consistent across both orderings, and inconsistent judgments are treated as ties.

The resulting pairwise preferences are converted into Elo ratings. We compute separate Elo ratings for fluency, hallucinations, and omissions. Because Elo scores can be sensitive to the order in which comparisons are processed, we repeat the computation over 1,000 random permutations of the comparison sequence and report the mean score across permutations.

\begin{figure*}[thb]
    \centering
    \begin{subfigure}{0.48\textwidth}
        \centering
        \includegraphics[width=\textwidth]{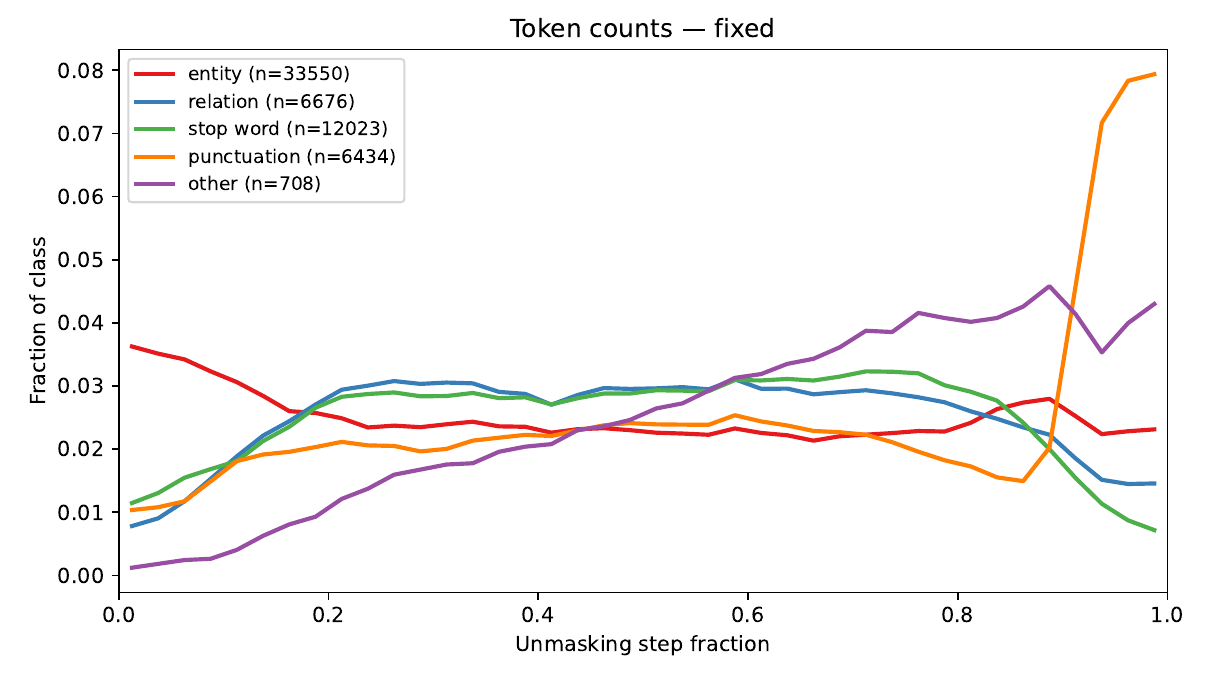}
        \caption{LLaMA}
        \label{fig:trajectory-small-llama}
    \end{subfigure}
    \hfill
    \begin{subfigure}{0.48\textwidth}
        \centering
        \includegraphics[width=\textwidth]{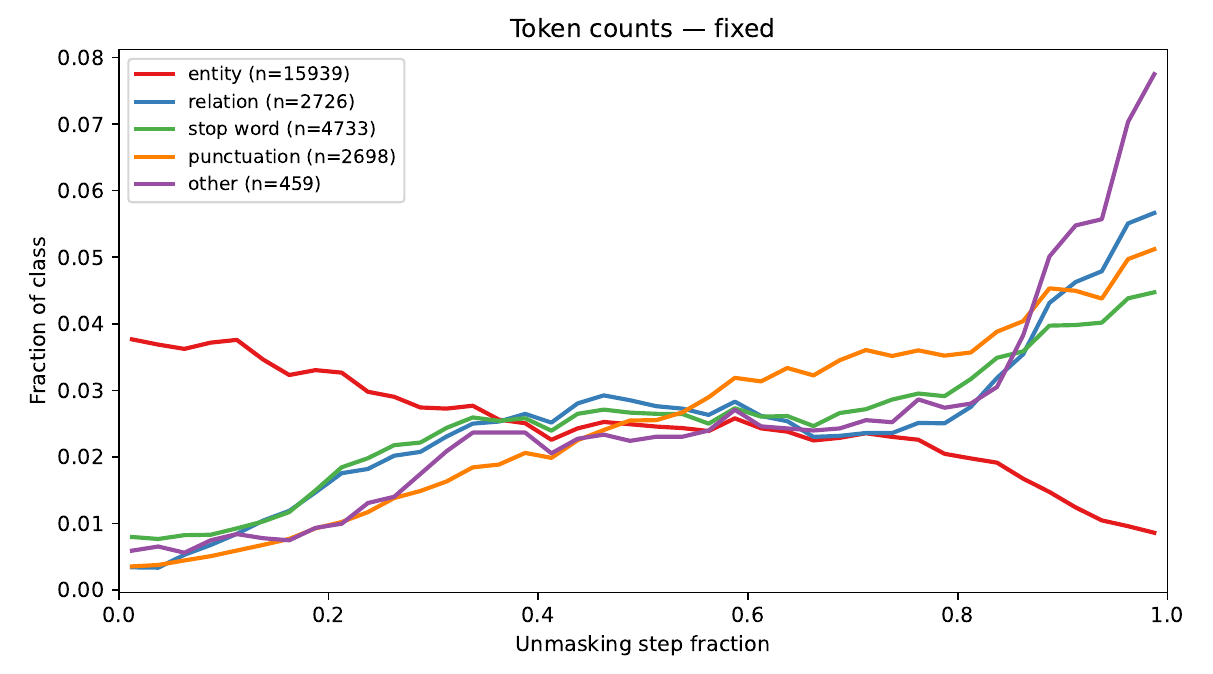}
        \caption{LLaDA}
        \label{fig:trajectory-small-llada}
    \end{subfigure}
    \caption{Trajectory comparison of LLaMA and LLaDA. An expanded comparison can be found in Figure~\ref{fig:trajectory-types}. LLaMA's trajectory is taken from its generation where it ``unmasks'' tokens left-to-right.}
    \label{fig:llama-llada-trajectory}
\end{figure*}

\section{Results and Analysis}

We organize our results around three questions: how MDLMs construct text from graph-structured inputs, whether their generation trajectories exhibit exploitable patterns, and whether they offer practical advantages over autoregressive LLMs. We first analyze the generation trajectories of LLaDA and Dream-7B, showing that $\lambda$-scaled structural decoding corrects the premature structural commitment induced by SFT. We then benchmark Graph-LLaDA against autoregressive and supervised baselines under both in-domain and out-of-domain conditions. Finally, we complement automatic metrics with human and LLM-based preference evaluations, which reveal that surface-level overlap metrics substantially understate MDLM generation quality and obscure meaningful differences in hallucination and omission rates.
 

\subsection{Analysis of Generation Trajectories}

\begin{table}[t]
\centering
\begin{tabular}{lcccc}
\toprule
 & \multicolumn{2}{c}{LLaMA 3.3} & \multicolumn{2}{c}{LLaDA} \\
\cmidrule(lr){2-3} \cmidrule(lr){4-5}
Type & Mean & Slope & Mean & Slope \\
\midrule
Entity      & 0.474 & $-0.308$ & 0.385 & $-1.352$ \\
Stop        & 0.501 & $-0.023$ & 0.576 & $+0.933$ \\
Relation    & 0.508 & $-0.105$ & 0.597 & $+1.181$ \\
Punct.      & 0.609 & $+1.309$ & 0.631 & $+1.591$ \\
Other       & 0.656 & $+1.876$ & 0.633 & $+1.620$ \\
\bottomrule
\end{tabular}
\caption{Token trajectory by token type. Trend refers to the slope of a line fit to its histogram density estimate.}
\label{tab:trajectory}
\vspace{-1em}
\end{table}

\paragraph{MDLMs exhibit strong entity-first generation.}
\label{sec:exp-traj-analysis}

As shown in Table~\ref{tab:trajectory} (and visually in Figure~\ref{fig:llama-llada-trajectory}), ARLMs left-to-right decoding process naturally follows the compositional structure of the target text, such as entity--relation--entity, with punctuation and boundary markers typically produced near the end of the sequence. In contrast, MDLMs are not constrained by this sequential decoding order. On both WebNLG and LAGRANGE, LLaDA and Dream consistently exhibit an entity-first unmasking pattern. Rather than constructing the output strictly from left to right, they preferentially recover high-confidence entity tokens at early denoising steps, and subsequently fill the remaining masked positions with relational, punctuation, and auxiliary content.

\paragraph{SFT induces premature structural commitment.} 
When compared with base LLaDA (Table~\ref{tab:llada_metrics_transposed}), SFT predicts structural tokens significantly earlier in the generation trajectory. Because the structure tokens are predicted early, it effectively fixes the output length before filling the content tokens. This leads to omitted or hallucinated outputs if the prematurely fixed length is too short or long. After SFT, the proportion of content tokens generated before the first period drop sharply from 60.9\% to 17.8\% and the mean and median first period tokens are generated significantly earlier (0.720 $\rightarrow$ 0.385). Furthermore, the number of period tokens doubles from 1.22 to 2.44 (due to the misplaced initial period) and the effective length of the sentence drops from 21.8 tokens to 20.3.

\begin{table}[t]
\centering
\small
\resizebox{\linewidth}{!}{
\begin{tabular}{lcccc}
\toprule
${\lambda}$ & {\small BLEU-4} & {\small METEOR} & {\small ROUGE\_L} & {\small CIDEr} \\
\midrule
\rowcolor{rowgray}
\multicolumn{5}{c}{\textit{LLaDA-8B\,$_{\text{SFT}}$}} \\
1.00 & 44.3 & 41.5 & 63.5 & 2.17 \\
0.75 & 45.6 & 41.8 & 65.1 & 2.38 \\
0.50 & \textbf{47.2} & \textbf{42.1} & \textbf{66.2} & \textbf{2.56} \\
0.25 & 45.0 & 41.8 & 65.5 & 2.41 \\
\midrule
\rowcolor{rowgray}
\multicolumn{5}{c}{\textit{Graph-LLaDA (Stage 1)}} \\
1.00 & 31.5 & 28.7 & 57.8 & 2.22 \\
0.50 & \textbf{49.1} & 37.6 & 65.5 & 2.86 \\
0.25 & 48.8 & \textbf{42.3} & \textbf{67.2} & \textbf{3.09} \\
\midrule
\rowcolor{rowgray}
\multicolumn{5}{c}{\textit{Graph-LLaDA (Stage 1 + Stage 2)}} \\
0.50 & 51.8 & 39.5 & 67.7 & 3.03 \\
0.25 & \textbf{55.4} & \textbf{43.4} & \textbf{69.5} & \textbf{3.37} \\
\bottomrule
\end{tabular}
}
\caption{Effect of $\lambda$-scaled structural decoding across configurations on WebNLG. Lower $\lambda$ delays the commitment of structural tokens. $\lambda=1.0$ corresponds to standard confidence-based remasking.}
\label{tab:alpha}
\vspace{-1em}
\end{table}

\paragraph{$\lambda$-scaled structural decoding mitigates premature commitment.}
Applying $\lambda$-scaled structural decoding effectively mitigates the tendency of SFT-tuned LLaDA to generate structural tokens at earlier stages of the trajectory by delaying structural commitment during generation. Table~\ref{tab:alpha_timing_analysis} illustrates how $\lambda$ controls the generation order of period and EOS tokens during unmasking. As $\lambda$ decreases, both period and EOS tokens are predicted progressively later in the generation trajectory. Meanwhile, the proportion of content tokens generated before the first period rises from 17.8\% to 89.9\%, indicating that lower $\lambda$ values encourage the model to resolve semantic content before committing to sentence structure. 

Interestingly, as shown in Figure~\ref{fig:trajectory-types}, the generation trajectory of $\lambda$-scaled MDLMs more closely resembles the sentence generation behavior observed in LLaMA models. This is largely because MDLMs use structure tokens as anchors for later content. Without them, MDLMs are less confident in predicting sparser and longer-range outputs.

\begin{table}[t]
\centering
\resizebox{\linewidth}{!}{
\begin{tabular}{lcccc}
\toprule
{Model} & {\small BLEU-4} & {\small METEOR} & {\small ROUGE\_L} & {\small CIDEr} \\
\midrule
\rowcolor{rowgray}
\multicolumn{5}{c}{\textit{Supervised Baselines}} \\
GAP (\textsc{lagrange})$\ddagger$        & 23.3 & 28.9 & 40.8 & 1.78 \\
MGSA$^\dagger$                  & \textbf{66.5} & 46.9 & \textbf{76.5} & -- \\
GAP (\textsc{WebNLG})$\ddagger$          & 66.3 & 46.9 & 76.3 & \textbf{4.65} \\
JointGT$^\dagger$               & 65.9 & \textbf{47.2} & 76.1 & -- \\
\midrule
\rowcolor{rowgray}
\multicolumn{5}{c}{\textit{Autoregressive Baselines}} \\
LLaMA 3.1-8B\,$_{\text{ZS}}$          & 44.5 & 37.9 & 61.2 & 2.64 \\
Qwen3-8B\,$_{\text{ZS}}$              & 48.4 & 40.5 & 63.8 & 2.90 \\
LLaMA 3.3-70B\,$_{\text{ZS}}$         & 51.4 & 41.0 & 66.9 & 3.15 \\
LLaMA 3.1-8B\,$_{\text{SFT}}$                & 57.2 & 42.6 & 68.8 & 3.38 \\
LLaMA 3.3-70B\,$_{\text{SFT}}$               & \textbf{57.6} & \textbf{42.8} & \textbf{69.4} & \textbf{3.44} \\
\midrule
\rowcolor{rowgray}
\multicolumn{5}{c}{\textit{Masked Diffusion Models}} \\
Dream-7B\,$_{\text{ZS}}$              & 23.4 & 25.5 & 56.9 & 2.23 \\
LLaDA-8B\,$_{\text{ZS}}$              & 42.2 & 36.1 & 62.3 & 2.50 \\
LLaDA-8B\,$_{\text{SFT}}$                    & 44.3 & 41.5 & 63.5 & 2.17 \\
LLaDA-8B\,$_{\text{SFT},\,\lambda{\texttt{=}}0.5}$   & 53.7 & \textbf{43.4} & 68.6 & 3.18 \\
Graph-LLaDA\,$_{\lambda{\texttt{=}}0.25}$ & \textbf{55.4} & \textbf{43.4} & \textbf{69.5} & \textbf{3.37} \\
\bottomrule
\end{tabular}
}
\caption{Main results on the WebNLG English test set. Best results within each model category are \textbf{bolded}. $\dagger$~denotes published results from~\citet{wang2024mgsa}. $\ddagger$~denotes our reproduced results. $_{\text{ZS}}$ indicates zero-shot.}
\label{tab:webnlg}
\end{table}

\begin{table}[t]
\centering
\resizebox{\linewidth}{!}{
\begin{tabular}{lcccc}
\toprule
{Model} & {\small BLEU-4} & {\small METEOR} & {\small ROUGE\_L} & {\small CIDEr} \\
\midrule
GAP (\textsc{WebNLG})                                 & 13.0 & 22.7 & 37.8 & 1.41 \\
Graph-LLaDA\,$_{\lambda{\texttt{=}}1.0}$      & 17.0 & 23.2 & \textbf{46.7} & 1.88 \\
LLaMA 3.1-8B                                  & 18.7 & 24.7 & 45.8 & \textbf{2.01} \\
LLaDA-8B\,$_{\lambda{\texttt{=}}1.0}$         & 15.4 & 24.9 & 42.1 & 1.44 \\
LLaDA-8B\,$_{\lambda{\texttt{=}}0.5}$         & 16.8 & \textbf{25.0} & 43.5 & 1.54 \\
LLaMA 3.3-70B                                 & 18.8 & \textbf{25.0} & 45.4 & 1.97 \\
Graph-LLaDA\,$_{\lambda{\texttt{=}}0.5}$      & \textbf{19.4} & \textbf{25.0} & {46.5} & \textbf{2.01} \\
\bottomrule
\end{tabular}
}
\caption{Results on the LAGRANGE test set. Best results among the evaluated models are \textbf{bolded}.}
\label{tab:main_results}
\vspace{-0.5em}
\end{table}

\subsection{Benchmark Comparison}
In the zero-shot setting, LLaDA-8B underperforms every autoregressive baseline including LLaMA~3.1-8B and Qwen3-8B\footnote{We focus subsequent discussion on LLaMA~3.1-8B as the primary autoregressive baseline, as Qwen3-8B's reasoning-oriented training makes standard LoRA fine-tuning difficult.}. SFT alone is insufficient to close this gap: LLaDA-8B$_\text{SFT}$ without $\lambda$-scaling loses 9.4 BLEU-4 points relative to the $\lambda$-scaled variant (Table~\ref{tab:webnlg}), directly implicating the premature structural commitment failure mode identified above. With both SFT and $\lambda=0.50$, LLaDA-8B reaches performance comparable to LLaMA~3.1-8B$_\text{SFT}$. Published supervised baselines substantially outperform all LLM/MDLM models in-domain (best: 66.5 vs.\ 57.6 BLEU-4).

\paragraph{Out-of-Domain Generalization.}
Out-of-domain evaluation on LAGRANGE (Table~\ref{tab:main_results}) reveals a markedly different picture. GAP exhibits the largest performance degradation across all models and is consistently outperformed by every LLM and MDLM. This indicates that small encoder-decoder supervised models learn dataset-specific stylistic patterns rather than generalizable graph-to-text mappings. LLM and MDLM models retain substantially more of their in-domain performance out of domain, confirming that scale and pretraining generalize more robustly than task-specific supervision alone.


\subsection{Human and LLM Evaluation}
Both humans and LLMs consistently assign high scores ($\geq$90\%) to Graph-LLaDA on fluency, omissions, and hallucinations (Table~\ref{tab:human-analysis}), indicating that its generation quality is considerably stronger than automatic metrics alone suggest. Agreement between annotators is very high, though Cohen's $\kappa$ remains low for most categories (Table~\ref{tab:agreement_kappa}). This is expected under skewed score distributions: when nearly all outputs receive the highest rating on a given dimension, there is little disagreement for $\kappa$ to measure, and the statistic collapses toward zero even when annotators are in near-perfect agreement. The high raw agreement rates are therefore more informative than $\kappa$ in this setting.

\paragraph{Pairwise LLM-based Evaluation.}
As shown in Tables~\ref{tab:elo_score_webnlg} and~\ref{tab:elo_score_lagrange}, GAP ranks among the worst-performing models under both in- and out-of-domain conditions despite its strong automatic metric scores, owing to hallucinations and omissions that lexical overlap metrics do not penalize. This further shows the inadequacy of surface-level metrics for graph-to-text evaluation.

\textbf{In-domain}, the dominant factor is model size: LLaMA~3.3-70B leads by a wide margin over all 8B-class models except Graph-LLaDA, which closes most of that gap despite being nearly nine times smaller. This suggests that the graph encoder provides a meaningful quality signal that partially compensates for the capacity disadvantage. \textbf{Out-of-domain}, the ranking shifts in favor of MDLMs. LLaDA-8B and Graph-LLaDA take the top two spots, with LLaMA~3.3-70B falling to third---a reversal that points to stronger generalization in masked diffusion models relative to their autoregressive counterparts, at least on this distribution. LLaDA's advantage is concentrated in fluency, while Graph-LLaDA's edge over base LLaDA is more pronounced on omissions, consistent with the graph encoder improving coverage of input triples.


\section{Related Work}
\label{sec:related-work}

\paragraph{Decoding Dynamics and Structural Token Behavior in MDLMs.}
Several concurrent works have observed related pathologies in MDLM decoding. \citet{kim2026earlydecisionsmatterproximity} note that LLaDA's SFT instruction model commits EOS tokens early before meaningful content is resolved, framing this as a length-control problem. \citet{yang2025tamingmaskeddiffusionlanguage} identify an ``EOS Trap'' in which EOS confidence is systematically higher than non-EOS tokens, and address it via block-wise decoding constraints. More broadly, \citet{wu2026stabilityweighteddecodingdiffusionlanguage} and \citet{shu2026deferredcommitmentdecodingdiffusion} propose training-free decoding strategies to mitigate premature commitment of unstable or high-uncertainty tokens across the trajectory. Our work differs in attributing the failure mode specifically to SFT, characterizing it via concrete trajectory statistics, and proposing $\lambda$-scaled structural decoding as a targeted fix.

\paragraph{Graph-to-text generation.}
Early systems relied on dedicated graph encoders~\cite{koncel-kedziorski-etal-2019-text}, but \citet{ribeiro-etal-2021-investigating} showed that linearizing triples and fine-tuning BART~\cite{lewis-etal-2020-bart} or T5~\cite{t5-base} reaches state-of-the-art on WebNLG without graph-specific inductive bias---at the cost of brittleness to meaning-preserving input permutations~\cite{hoyle-etal-2021-promoting}. A second line of work re-injects graph structure into a pretrained backbone: JointGT~\cite{ke-etal-2021-jointgt} adds structure-aware aggregation and graph--text alignment objectives, GAP~\cite{colas-etal-2022-gap} uses only graph-aware attention masks, and MGSA~\cite{wang2024mgsa} combines entity- and word-level structural attention. More recently, LAGRANGE~\cite{mousavi-etal-2024-construction} introduces a much larger Wikipedia/Wikidata-aligned KG--text corpus, constructed under a cyclic-equivalence criterion to improve graph--text faithfulness; we use it in Section~\ref{sec:exp-traj-analysis} as an out-of-distribution test set to probe whether WebNLG-trained models actually generalize. Our work follows the structural-encoding tradition but is the first to pair it with a masked diffusion decoder.

\paragraph{Masked diffusion language models.}
Discrete diffusion for text traces back to D3PM~\cite{austin2021d3pm}, whose absorbing-state variant connects diffusion to masked language modeling. MDLM~\cite{sahoo2024mdlm} reduces the objective to a mixture of masked-LM losses competitive with autoregressive perplexity, and SEDD~\cite{lou2024sedd} surpasses GPT-2 via a score-entropy reformulation. LLaDA~\cite{nie2026large} is the first 8B masked diffusion LLM trained from scratch and is competitive with LLaMA-3 8B on instruction following; Dream-7B~\cite{ye2025dream} reaches comparable performance by initializing from an autoregressive checkpoint with context-adaptive noise rescheduling. Closer to our setting, \citet{kim2025any} propose any-order, flexible-length masked diffusion as a training-time fix for fixed-length generation; our $\lambda$-scaled structural decoding addresses a related symptom---premature commitment to length-fixing structural tokens---but requires no retraining, and we return to this comparison in the conclusion as the natural principled alternative to our inference-time intervention.

\paragraph{Graph-augmented LLMs.}
Several recent systems couple a GNN encoder with a pretrained autoregressive LLM. G-Retriever~\cite{he2024gretriever} extracts query-relevant subgraphs and feeds them to a frozen LLM via a \emph{single} pooled graph token, while GraphGPT~\cite{tang2024graphgpt} uses a per-node sequence of graph tokens delimited by start/end markers; \citet{grover2026one} show that G-Retriever's single-token pooling creates an information bottleneck that more graph tokens recover. Graph-LLaDA inherits GraphGPT's per-node injection pattern but combines it with a masked diffusion decoder. To the best of our knowledge, no prior work has paired an MDLM with a graph encoder for \emph{any} task, and no discrete-diffusion language model of any kind has been applied to KG-to-text generation: Graph-LLaDA is novel on both axes.

\section{Conclusion}
We presented the first systematic study of masked diffusion language models for graph-to-text generation. By recording the full unmasking trajectories of LLaDA and Dream-7B on WebNLG and LAGRANGE, we showed that masked diffusion naturally adopts a content-first decoding order---entities first, then relations and function words, with structural tokens last---without any explicit decoding-order supervision. 

In analyzing the same trajectories under supervised fine-tuning, we identified a previously undocumented failure mode of SFT-tuned MDLMs: premature commitment to sentence-ending tokens, which fixes output length and segmentation before substantive content has been resolved. We addressed this with $\lambda$-scaled structural decoding, a training-free inference-time fix that recovers up to $+9.4$ BLEU-4. Building on these findings, we proposed Graph-LLaDA, which integrates a Graph Transformer encoder into LLaDA via a two-stage training pipeline. Under pairwise LLM-judge evaluation, Graph-LLaDA closes most of the in-domain gap to LLaMA~3.3-70B and ranks second out-of-domain on LAGRANGE---ahead of LLaMA~3.3. 



\section*{Limitations}

\paragraph{Single-run experiments.}
Due to the large number of experiments run and the known computational inefficiencies of LLaDA, we only run our experiments a single time. Because LLaDA uses confidence-based generation, we expect outputs to remain consistent across runs and our prior experience supports this, though we did not validate it in this work. 

\paragraph{Single MDLM backbone for SFT and $\lambda$-scaling.}
Our SFT, $\lambda$-scaling, and Graph-LLaDA experiments all use LLaDA-8B as the backbone. While we replicate the entity-first trajectory finding on Dream-7B (Appendix~\ref{app:dream}), the SFT failure mode and the $\lambda$-scaling fix have only been validated on LLaDA, and we have not tested them on Dream-7B, LLaDA-MoE, or LLaDA~2. We expect the findings to generalize to other MDLMs but do not claim this is established.

\paragraph{Single-token-per-step decoding.}
All experiments use single-token-per-step generation, since multi-token decoding degrades quality substantially in our setting. Combining $\lambda$-scaling with recent multi-token decoding strategies is left to future work.

\paragraph{English graph-to-text only.}
Our evaluation is limited to two graph-to-text benchmarks (WebNLG and LAGRANGE) in English; we do not investigate multilingual generation or knowledge-graph question answering settings.

\paragraph{LLM-judge Elo bias.}
The LLM-judge Elo evaluation, while controlled for positional bias and aggregated over many permutations, inherits whatever biases the judge models hold. We treat it as a complement to, not a replacement for, the (smaller) human evaluation.

\paragraph{Rule-based token classification.}
Our token-classification procedure for trajectory analysis is rule-based and has a small residual \textit{other} category that we exclude from the categorical means. This residual is concentrated late in the trajectory and is reported in Table~\ref{tab:trajectory}.

\bibliography{custom}

\appendix

\section{Methodology Details}
\label{app:methodology}

\subsection{Implementation Details}
\label{app:multi-token-generation}

For LLaDA SFT, we fine-tune with LoRA~\cite{hu2022lora} ($r=16$, $\alpha=32$) for 5 epochs using random-token masking. For Graph-LLaDA, the GNN encoder uses a Graph Transformer with a hidden dimension of 512 (See Table~\ref{tab:dim_ablation}), and the projector is a 2-layer MLP. Stage~1 training (frozen LLaDA, GNN and projector only) runs for 5 epochs; Stage~2 (joint fine-tuning with attention-only LoRA) runs for 3 epochs. During inference, we use iterative unmasking with confidence-based token selection. Unless otherwise noted, we apply $\lambda$ scaling ($\lambda = 0.50$) to structural tokens and set the generation length to 128 tokens. Qwen3-8B is evaluated with chain-of-thought reasoning enabled, as disabling it substantially degrades performance (BLEU-4 drops from 48.35 to 25.10).

All training and inference was done on nodes with 1x--8x H100 GPUs. Only LLaMA 3.3 required 2 or more GPUs, all LLaDA models can fit on a single H100. Claude Code, OpenAI Codex, and/or Github Copilot were used in the development of this work.

While masked diffusion models can generate multiple tokens per step, in this paper we study single-token generation only, as multi-token generation leads to significant performance degradation. Additionally, many recent works~\cite{zhang2025survey} have explored different strategies for improving multi-token generation, thus any findings related to standard multi-token generation may not be applicable to real-world implementations.

\subsection{Token Classification Details}
\label{app:token-classification}

This section expands the priority-based token classification procedure summarized in Section~\ref{sec:methodology}. Each generated token is assigned to exactly one of five categories (\textit{entity}, \textit{relation}, \textit{stop word}, \textit{punctuation}, \textit{other}) by the first matching rule.

\paragraph{Surface-form matching against the input triples.}
For each triple, subject and object names are normalized (underscores replaced with spaces, parenthetical suffixes stripped) and augmented with common format variants (numeric comma formatting, date expansions). Predicate names are similarly normalized and split on camelCase and underscore boundaries into individual content words. A case-insensitive search over the decoded output text marks spans that match an entity or relation surface form, with entity matches taking priority over relation matches when spans overlap.

\paragraph{Linguistic classification with spaCy.}
For tokens not resolved by surface-form matching, we apply spaCy-based (en\_core\_web\_sm) linguistic classification. Tokens whose named-entity label falls within a standard set (\textsc{person}, \textsc{org}, \textsc{gpe}, \textsc{date}, etc.) are labeled as entities; proper nouns and numerals are also labeled as entities. Verbs are labeled as relations, and nouns and adjectives that fuzzy-match a relation content word are likewise labeled as relations. Punctuation tokens are identified by spaCy's \texttt{is\_punct} flag, and function words by \texttt{is\_stop} or membership in a closed set of part-of-speech tags (\textsc{det}, \textsc{adp}, \textsc{aux}, \textsc{pron}, \textsc{cconj}, \textsc{sconj}, \textsc{part}).

\paragraph{Fallback lexicon.}
As a final fallback, tokens still labeled as \textit{other} are checked against a conservative manually-built lexicon of common relation realizations (e.g., \emph{married}, \emph{band}, \emph{city}) and reclassified as relations if matched. End-of-sequence and end-of-turn special tokens are excluded from the analysis throughout.

\subsection{Prompt Templates}
\label{app:prompts}

We use a single instruction template across all models. The template is wrapped in each model's chat format using \texttt{tokenizer.apply\_chat\_template} with \texttt{add\_generation\_prompt=True}. We deliberately did not tune prompts per model so that performance differences reflect model capability rather than prompt engineering.

\paragraph{Shared instruction (LLaDA zero-shot / SFT, LLaMA, Qwen).}
The triples list is interpolated as one bullet per triple, using the WebNLG \texttt{subject | predicate | object} format verbatim.

\begin{quote}\small\ttfamily
You are given a set of knowledge graph triples. Generate a single natural-language sentence that describes the information in the triples.\\[2pt]
Requirements:\\
- The sentence must be fully consistent with the triples.\\
- The sentence must cover all facts in the triples.\\
- The sentence must not add any facts not present in the triples.\\
- Only output the sentence and nothing else.\\[2pt]
Triples:\\
\{list\_of\_triples\} \\
\end{quote}

\noindent For Qwen3-8B the reasoning mode is disabled by appending \texttt{/no\_think} after the rendered chat template, following the model's documented switch~\cite{qwen3technicalreport}. LLaMA and Dream use the rendered template unchanged.

\paragraph{Graph-LLaDA.}
For Graph-LLaDA the user message contains both a bracketed rendering of the triples (for lexical grounding) and a single \texttt{<graph>} marker whose embedding is replaced at the input-embedding layer by the projected GNN node sequence (Section~\ref{sec:graph-llada}). The expanded token stream the model actually sees is \texttt{<g\_start> <g\_patch> ... <g\_patch> <g\_end>}, with one \texttt{<g\_patch>} per graph node.

\begin{quote}\small\ttfamily
You are given a knowledge graph. Generate a single natural-language sentence that describes the information in the graph.\\[2pt]
Requirements:\\
- The sentence must be fully consistent with the graph.\\
- The sentence must cover all facts in the graph.\\
- The sentence must not add any facts not present in the graph.\\
- Reuse the entity and relation names from the triples when possible.\\
- Only output the sentence and nothing else.\\[2pt]
Triples:\\
\{list\_of\_triples\} \\
Graph tokens: <graph>
\end{quote}

\section{Additional Results}
\label{app:additional-results}

\begin{table}[thb]
\centering

\begin{tabular}{ccccc}
\toprule
\textbf{Dimension} & \textbf{BLEU-4} & \textbf{METEOR} & \textbf{Brevity} \\
\midrule
256 & 50.30 & 38.54 & 0.885 \\
\textbf{512} & \textbf{53.61} & \textbf{41.18} & \textbf{1.012} \\
1024 & 47.50 & 36.72 & 0.837 \\
2048 & 47.94 & 36.90 & 0.858 \\
4096 & 45.07 & 35.46 & 0.789 \\
\bottomrule
\end{tabular}
\caption{Performance comparison of different hidden dimension sizes for graph encoder.}
\label{tab:dim_ablation}
\end{table}

\subsection{Ablation Studies}
\label{app:ablation-study}
 
\paragraph{Stage~1 LoRA Configuration.} Table~\ref{tab:dim_ablation} reports the effect of GNN hidden dimension on Stage~1 performance (frozen LLaDA, 5 epochs). Performance is non-monotonic, peaking at dimension 512 (53.61 BLEU-4) and declining at higher dimensions. At 5 epochs, dimension 512 already matches dimension 256 trained for 8 epochs (53.56 BLEU-4). The gap widens with graph complexity: at 5 triples, dimension 512 leads by 7.34 BLEU-4. Larger dimensions (1024--4096) consistently underperform, likely because the high-dimensional GNN representations do not project effectively into LLaDA's embedding space.
 
\paragraph{Stage~2 LoRA Configuration.}
Starting from a dimension-512 Stage~1 model, we compare attention-only LoRA against all-linear LoRA for Stage~2 fine-tuning. Attention-only LoRA provides a modest gain (+0.52 BLEU-4 to 54.13), while all-linear LoRA slightly hurts overall performance ($-$0.77 BLEU-4 to 52.84). The additional capacity from adapting MLP layers introduces instability on complex graphs: 7-triple BLEU-4 drops from 41.17 to 36.10. The WebNLG training set (35K examples) appears too small to benefit from the larger parameter budget of all-linear LoRA (21M vs 12.6M trainable parameters).


\begin{table*}[thb]
\centering
\small
\begin{tabular}{lccccccc|c}
\toprule
\textbf{Model} & \textbf{1} & \textbf{2} & \textbf{3} & \textbf{4} & \textbf{5} & \textbf{6} & \textbf{7} & \textbf{Spread} \\
\midrule
\textit{$N$ per bucket} & 369 & 349 & 350 & 305 & 213 & 114 & 79 & --- \\
\midrule
LLaDA-8B (zero-shot)                 & 62.41 & 55.21 & 44.23 & 36.12 & 32.53 & 37.80 & 40.06 & 29.88 \\
LLaMA~3.1-8B (zero-shot)             & 58.97 & 49.90 & 46.34 & 42.37 & 41.48 & 38.29 & 40.88 & 20.68 \\
Qwen3-8B (zero-shot)                 & 59.05 & 55.18 & 50.67 & 46.77 & 44.96 & 41.64 & 43.59 & 17.41 \\
LLaMA~3.3-70B (zero-shot)            & {64.38} & 58.76 & 54.71 & 50.37 & 47.34 & 43.55 & 43.42 & 20.96 \\
LLaDA-8B (SFT, $\lambda{=}0.5$) & 63.65 & 57.14 & 54.19 & 52.62 & \textbf{51.94} & \textbf{50.83} & 50.17 & \textbf{13.48} \\
Graph-LLaDA ($\lambda{=}0.25$)       & \textbf{68.42} & \textbf{60.30} & \textbf{58.04} & \textbf{55.78} & 51.76 & 47.70 & \textbf{51.70} & 16.72 \\
\bottomrule
\end{tabular}
\caption{BLEU-4 on the WebNLG English test set, broken down by the number of input triples. \textit{Spread} is the difference between the best and worst per-bucket scores for each model and measures robustness to graph complexity. Best in each column is \textbf{bolded}.}
\label{tab:per_triple}
\end{table*}

\subsection{Per-Triple Performance Breakdown}
\label{app:per-triple}
As shown in Table~\ref{tab:per_triple}, BLEU-4 consistently decreases as the number of input triples increases across all models. The degradation is most pronounced for zero-shot LLaDA-8B, which drops sharply from 62.41 to 32.53, indicating its limited ability to handle larger graphs. SFT mitigates this: SFT LLaDA-8B achieves the smallest performance spread of 13.48 across triple counts. Graph-LLaDA ($\lambda=0.25$) obtains the best BLEU-4 scores in buckets 1--4 and 7, while maintaining a spread of 16.72, suggesting that the graph encoder improves both overall quality and robustness to graph complexity.




\begin{figure*}[thb]
    \centering
    \begin{subfigure}{0.9\textwidth}
        \centering
        \includegraphics[width=\textwidth]{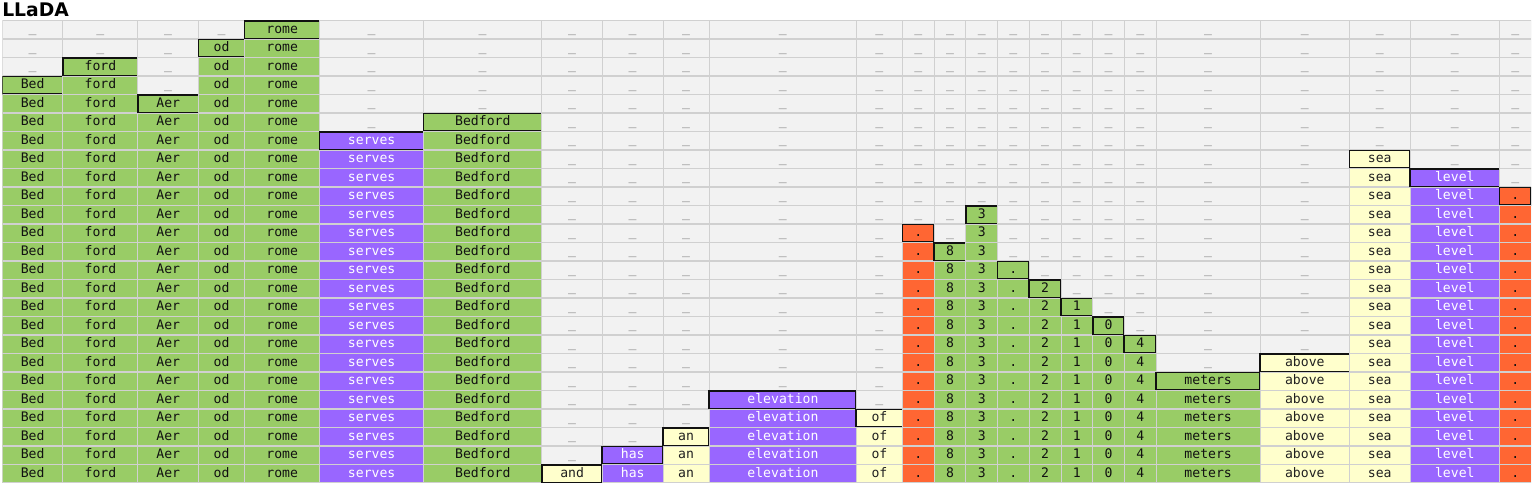}
        \caption{LLaDA: Bedford Aerodrome serves Bedford and has an elevation of 83.2104 meters above sea level.}
        \label{fig:llada_text}
    \end{subfigure}
    \hfill
    \begin{subfigure}{0.9\textwidth}
        \centering
        \includegraphics[width=\textwidth]{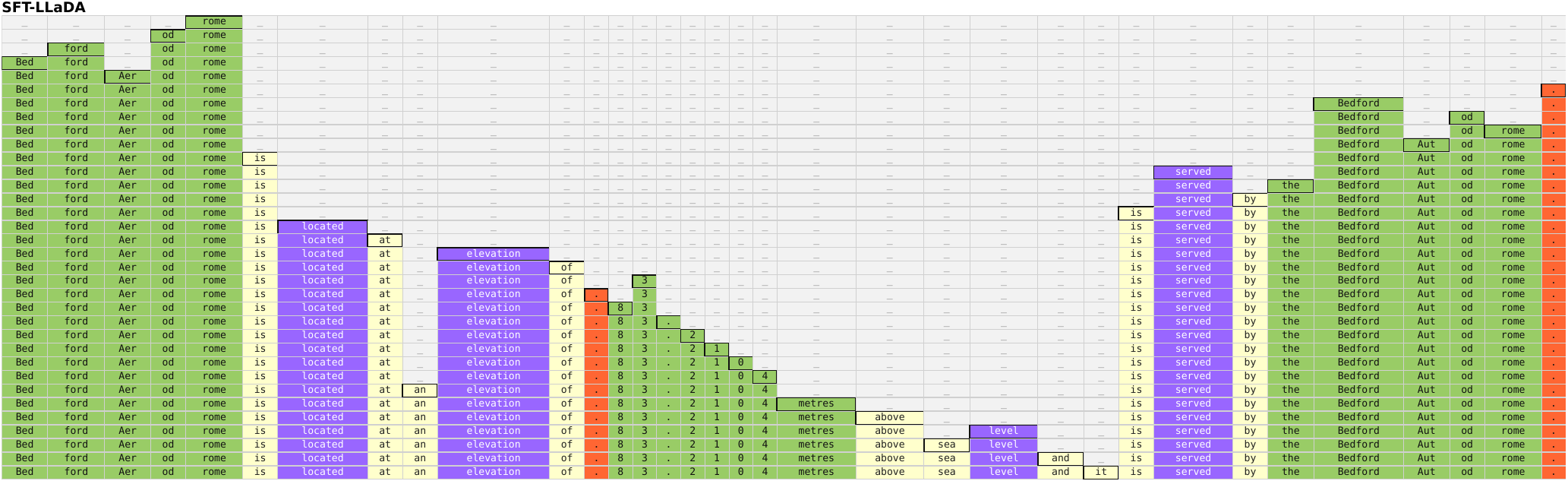}
        \caption{SFT-LLaDA: Bedford Aerodrome is located at an elevation of 83.2104 metres above sea level and it is served by the Bedford Autodrome.}
        \label{fig:sft_text}
    \end{subfigure}
    \hfill
    \begin{subfigure}{0.9\textwidth}
        \centering
        \includegraphics[width=\textwidth]{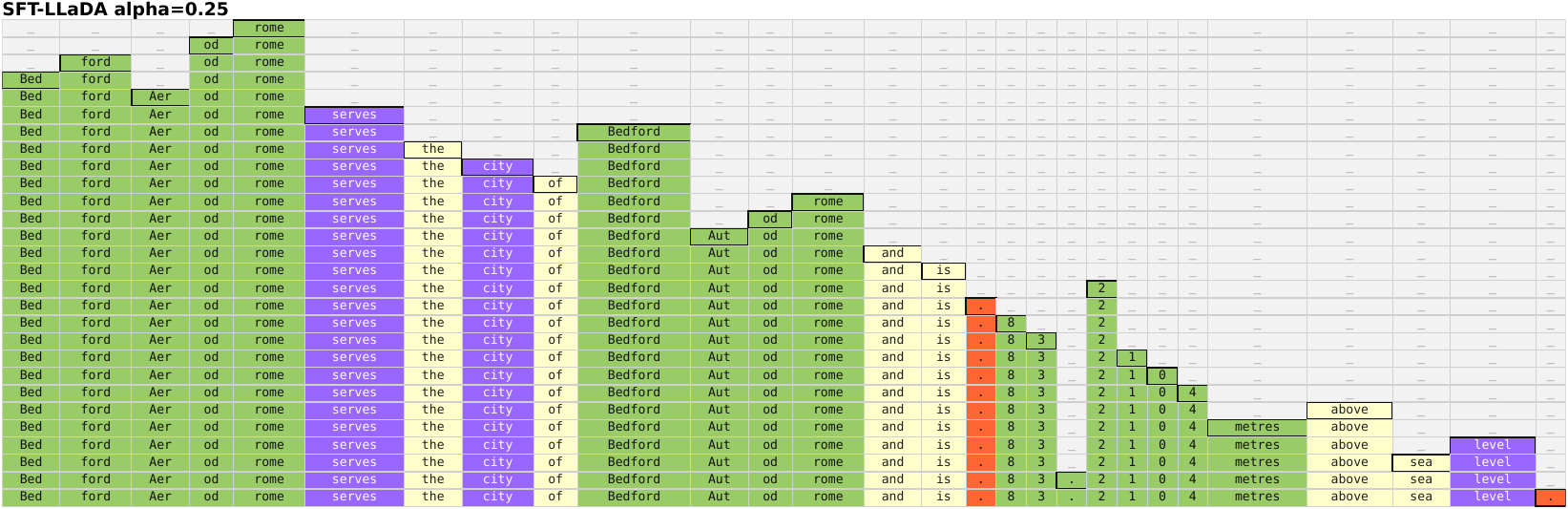}
        \caption{SFT-LLaDA ($\lambda=0.25$): Bedford Aerodrome serves the city of Bedford Autodrome and is 83.2104 metres above sea level.}
        \label{fig:sft_llada_text}
    \end{subfigure}
    \caption{Token unmasking trajectory of MDLMs on a WebNLG sample with input triples \texttt{(Bedford\_Aerodrome, cityServed,  Bedford\_Autodrome)} and \texttt{(Bedford\_Aerodrome , elevationAboveTheSeaLevel, 83.2104)}.}
    \label{fig:mdlms_output}
\end{figure*}
\begin{table*}[thb]
\centering
\footnotesize

\begin{tabular}{lcccc}
\toprule
\textbf{$lambda$} & 
\textbf{First "." step (mean)} & 
\textbf{Content before first "."} & 
\textbf{"." mean step} & 
\textbf{EOS mean step} \\
\midrule
1.00 & 0.385 & 17.8\% & 0.522 & 0.344 \\
0.75 & 0.414 & 37.6\% & 0.570 & 0.430 \\
0.50 & 0.436 & 64.9\% & 0.636 & 0.579 \\
0.25 & 0.607 & 89.9\% & 0.774 & 0.639 \\
\bottomrule
\end{tabular}

\caption{Timing analysis across different $\lambda$ values, showing when punctuation and EOS tokens are unmasked relative to content tokens during generation.}
\label{tab:alpha_timing_analysis}
\end{table*}

\subsection{Trajectory Visualizations and Detailed Metrics}
\label{app:trajectory-visuals}
\begin{figure*}[thb]
    \centering
    \begin{subfigure}{0.48\textwidth}
        \centering
        \includegraphics[width=\textwidth]{figures/trajectories/trajectory-llama.pdf}
        \caption{LLaMA}
        \label{fig:trajectory-llama}
    \end{subfigure}
    \hfill
    \begin{subfigure}{0.48\textwidth}
        \centering
        \includegraphics[width=\textwidth]{figures/trajectories/trajectory-llada.pdf}
        \caption{LLaDA}
        \label{fig:trajectory-llada}
    \end{subfigure}
    \begin{subfigure}{0.48\textwidth}
        \centering
        \includegraphics[width=\textwidth]{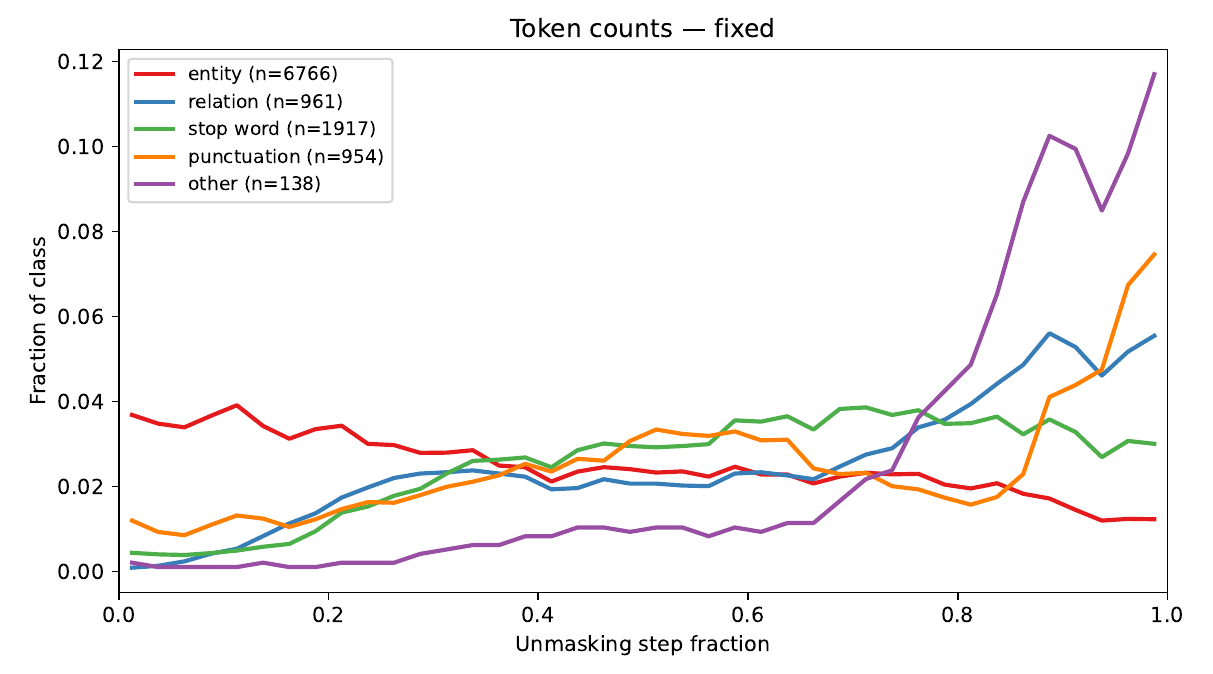}
        \caption{Dream-7B}
        \label{fig:trajectory-dream}
    \end{subfigure}
    \begin{subfigure}{0.48\textwidth}
        \centering
        \includegraphics[width=\textwidth]{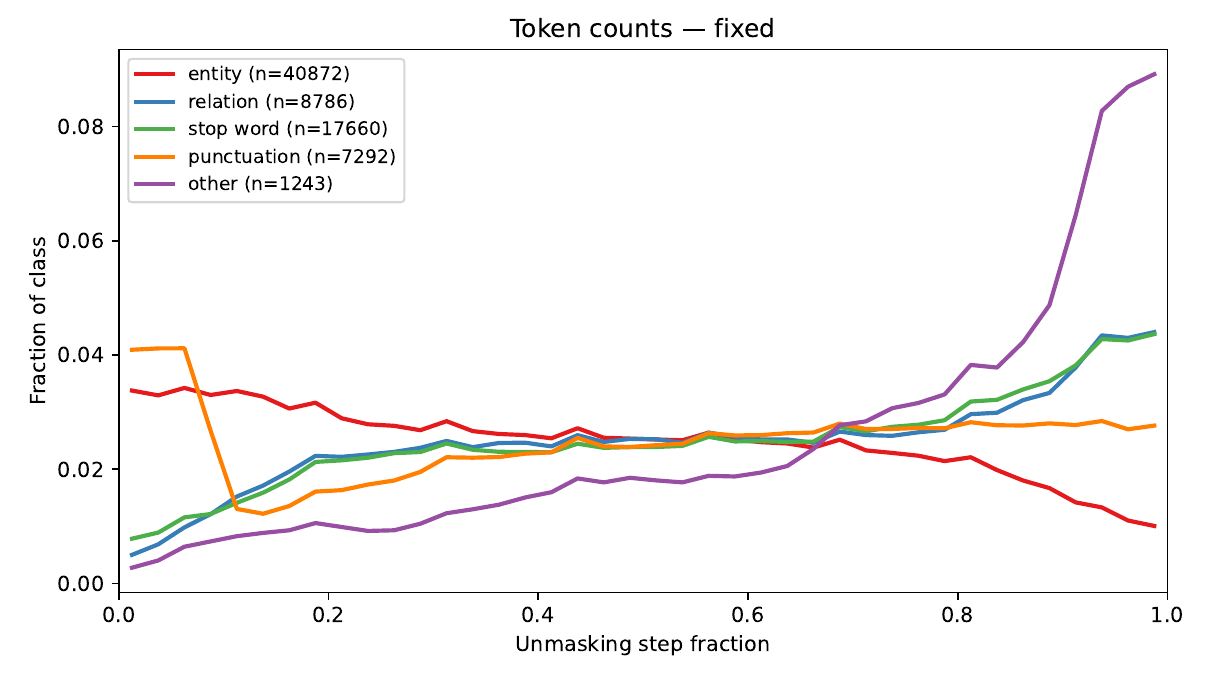}
        \caption{LLaDA SFT}
        \label{fig:trajectory-llada-sft}
    \end{subfigure}
    \hfill
    \begin{subfigure}{0.48\textwidth}
        \centering
        \includegraphics[width=\textwidth]{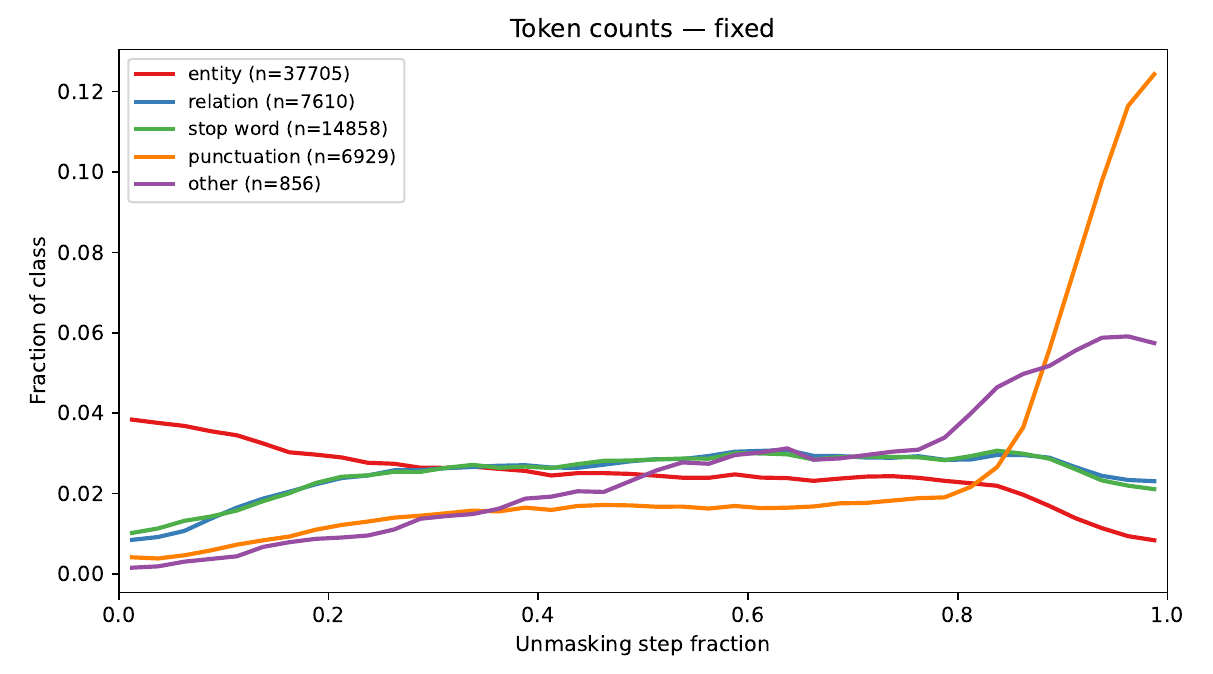}
        \caption{LLaDA SFT ($\lambda=0.5$)}
        \label{fig:trajectory-llada-sft-lambda-50}
    \end{subfigure}
    \hfill
    \begin{subfigure}{0.48\textwidth}
        \centering
        \includegraphics[width=\textwidth]{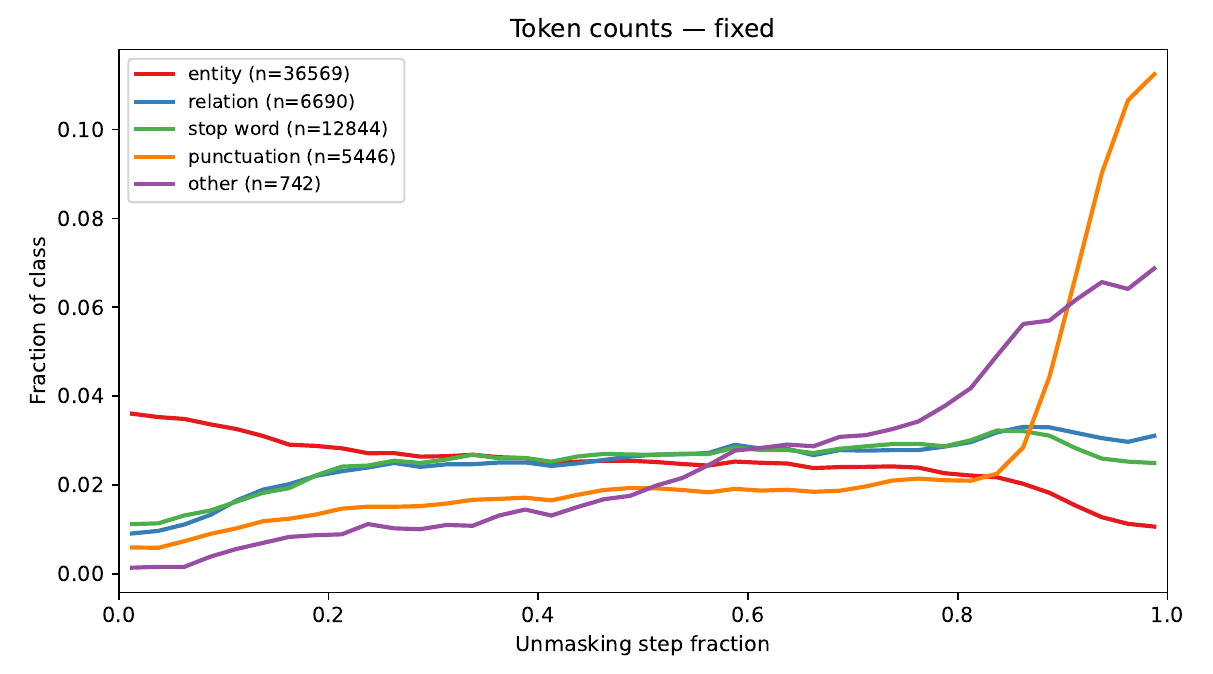}
        \caption{Graph-LLaDA ($\lambda=0.5$)}
        \label{fig:trajectory-graph-llada-lambda-50}
    \end{subfigure}
    \caption{Trajectory comparison of classified token types for LLaDA and autoregressive models.}
    \label{fig:trajectory-types}
\end{figure*}






\begin{table}[t]
\centering
\footnotesize
\small
\begin{tabular}{lccc}
\toprule
\textbf{Metric} & \textbf{LLaDA} & \textbf{LLaDA(SFT)} & \textbf{Delta} \\
\midrule
Mean First "."           & 0.720 & 0.385 & -0.335 \\
Median First "."        & 0.891 & 0.484 & -0.407 \\
\midrule
Content Before "."      & 60.9\% & 17.8\% & -43.1\% \\
\midrule
Periods / Sample        & 1.22 & 2.44 & +1.22 \\
EOS / Sample            & 37.2 & 21.4 & -15.8 \\
Effective length        & 21.8 & 20.3 & -1.5 \\
\bottomrule
\end{tabular}
\caption{Comparison between Base LLaDA and SFT LLaDA across punctuation and generation metrics, showing when structural tokens (e.g., ``.'', EOS, and EOT) are unmasked relative to content tokens in generation trajectories.}
\label{tab:llada_metrics_transposed}
\end{table}

To better understand why supervised fine-tuning yields smaller performance gains for LLaDA than for other models, we further analyze the graph-to-text generation process, as illustrated in Figure~\ref{fig:trajectory-types} and Figure~\ref{fig:mdlms_output}. This subsection collects the trajectory figures and supporting tables that accompany the trajectory analysis of Section~\ref{sec:exp-traj-analysis}. Figure~\ref{fig:trajectory-types} shows aggregated generation-step distributions by token category across LLaMA, base LLaDA, SFT LLaDA, SFT LLaDA with $\lambda$-scaling, and Graph-LLaDA. Figure~\ref{fig:mdlms_output} provides a direct base-vs-SFT comparison. We quantify this phenomenon in Table~\ref{tab:llada_metrics_transposed}. Tables~\ref{tab:llada_metrics_transposed} and~\ref{tab:alpha_timing_analysis} report the first-period and content-before-period statistics that underpin the SFT failure-mode discussion and the $\lambda$-scaling recovery curve.

\subsection{Dream-7B Trajectory Analysis}
\label{app:dream}

\todo[color=red]{Put Dream results in a table and reference it.}

We replicate the trajectory analysis on Dream-7B's WebNLG predictions using the procedure of Section~\ref{sec:methodology}, with Dream's own tokenizer. Mean unmasking steps are reported alongside LLaDA in Table~\ref{tab:trajectory}; we summarize the comparison here.

\paragraph{The entity-first ordering generalizes.} Dream commits entities at mean step $0.423$ with slope $-0.90$, closely matching LLaDA's $0.385$ ($-1.35$) on WebNLG and $0.427$ ($-0.87$) on LAGRANGE. Entities are clearly separated from every other category in both backbones, supporting the claim that content-first decoding is a property of masked diffusion rather than of any particular MDLM.

\paragraph{Fine-grained structural token ordering diverges.} In LLaDA, punctuation is the latest-committed category (mean $0.631$), producing a clean three-tier ordering (entities, relations/stop words, then punctuation). In Dream, punctuation ($0.616$) is interleaved with stop words ($0.613$) and slightly precedes relations ($0.655$), collapsing to a two-tier structure (entities, then everything else). 

We do not interpret the Dream-vs-LLaDA divergence as evidence about which architecture is ``correct''---it may instead reflect differences in pretraining objective, instruction tuning, or output-length calibration. We did not run Dream SFT or $\lambda$-scaling experiments and so cannot directly compare its post-SFT failure mode to LLaDA's.

\subsection{Generation Examples}
\label{app:examples}
\begin{figure*}[thb]
\small
\centering

\scalebox{0.8}{%
\begin{minipage}{\textwidth}

\begin{tcolorbox}[
colback=gray!10,
colframe=blue!70!black,
title=Example 1. Graph-to-Text on WebNLG,
fonttitle=\bfseries,
sharp corners,
boxrule=1pt
]


\textbf{Knowledge Graph}: 
\texttt{
\{(Hypermarcas | location | São\_Paulo), (Brazil | areaTotal | 8514837.14  (square kilometres)),(Hypermarcas | location | Brazil), (Hypermarcas | industry | Pharmaceuticals)\}
}
\vspace{0.5em}

\textbf{LLaDA}:
\texttt{
Hypermarcas is a pharmaceutical company located in São Paulo, Brazil.
}

\vspace{0.5em}
\textbf{LLaDA-8B\,$_{\lambda{\texttt{=}}0.5}$}:
\texttt{Hypermarcas is a pharmaceutical company located in São Paulo, Brazil. Hypermarcas is located in Brazil, which has a total area of 8514837.14 square kilometres. Brazil has a total area of 8514837.14 square kilometres.
}

\vspace{0.5em}
\textbf{Graph-LLaDA\,$_{\lambda{\texttt{=}}1.0}$ }:
\texttt{Hypermarcas is a pharmaceutical company located in São Paulo, Brazil.
}

\vspace{0.5em}
\textbf{Graph-LLaDA\,$_{\lambda{\texttt{=}}0.25}$ }:
\texttt{Hypermarcas is a pharmaceutical company located in São Paulo, Brazil which has a total area of 8514837.14 square kilometres.
}
\end{tcolorbox}

\begin{tcolorbox}[
colback=gray!10,
colframe=blue!70!black,
title=Example 2 for Graph-to-Text on WebNLG,
fonttitle=\bfseries,
sharp corners,
boxrule=1pt
]


\textbf{Knowledge Graph}: 
\texttt{
\{(Super\_Capers | writer | Ray\_Griggs\_(director)), (Ray\_Griggs\_(director) | birthYear | 1974), (Super\_Capers | starring | Michael\_Rooker), (Super\_Capers | starring | Tom\_Sizemore)\}
}
\vspace{0.5em}

\textbf{LLaDA}:
\texttt{
Ray Griggs directed Super Capers, starring Michael Rooker and Tom Sizemore.
}

\vspace{0.5em}
\textbf{LLaDA-8B\,$_{\lambda{\texttt{=}}0.5}$}:
\texttt{Ray Griggs, who was born in 1974, is the director of the film `Super Capers'. He is also the writer of the film `Super Capers' which stars Tom Sizemore. The film 'Super Capers' stars Michael Rocker and Tom Sizemore.
}

\vspace{0.5em}
\textbf{Graph-LLaDA\,$_{\lambda{\texttt{=}}1.0}$ }:
\texttt{Super Capers stars Michael Goulding and Tomizemore.
}

\vspace{0.5em}
\textbf{Graph-LLaDA\,$_{\lambda{\texttt{=}}0.25}$ }:
\texttt{Super Capers was written by Ray Griggs who was born in 1974. It starred Michael Reding and Tom Sizemore.
}

\end{tcolorbox}

\begin{tcolorbox}[
colback=gray!10,
colframe=blue!70!black,
title=Example 3 for Graph-to-Text on LAGRANGE,
fonttitle=\bfseries,
sharp corners,
boxrule=1pt
]


\textbf{Knowledge Graph}: 
\texttt{
\{(English\_Without\_Tears | director | Harold\_French), (Terence\_Rattigan | deathYear | 1977-01-01), (Harold\_French | activeYearsStartYear | 1920-01-01), (English\_Without\_Tears | writer | Anatole\_de\_Grunwald), (English\_Without\_Tears | writer | Terence\_Rattigan)\}
}
\vspace{0.5em}

\textbf{LLaDA}:
\texttt{
"English Without Tears" is a film directed by Harold French, who has been active in the film industry since 1920, and the film has been written by multiple writers, including Anatole de Grunwald and Terence Rattigan, who passed away in 1977.
}

\vspace{0.5em}
\textbf{LLaDA-8B\,$_{\lambda{\texttt{=}}0.5}$}:
\texttt{Harold French, the director of English Without Tears, began his career on January 1, 1920. He directed the film "English Without Tears" which was written by Anatole de Grunwald and Terence Rattigan, who died on January 1, 1977.
}

\vspace{0.5em}
\textbf{Graph-LLaDA\,$_{\lambda{\texttt{=}}1.0}$ }:
\texttt{Terence Ratigan, who wrote the play English Without Tears died on
}

\vspace{0.5em}
\textbf{Graph-LLaDA\,$_{\lambda{\texttt{=}}0.25}$ }:
\texttt{Terence Rattigan and Anatole de Grunwald are the writers of English Without Tears, which was directed by Harold French. Terence Rattigan died on January 1, 1977 and Harold French began his career on January 1, 1920.
}

\end{tcolorbox}

\end{minipage}%
}

\caption{Generation examples from WebNLG (Examples 1--2) and LAGRANGE (Example 3) comparing base LLaDA, SFT LLaDA, and Graph-LLaDA under varying $\lambda$ settings.}

\label{fig:example}
\end{figure*}

Representative examples are shown in Figure~\ref{fig:example}. Base LLaDA tends to produce fluent outputs with few hallucinations but omits facts from the input graph. SFT alleviates omissions but can introduce repetition. Graph-LLaDA without $\lambda$-scaling is less stable in controlling output length, leading to truncated or hallucinated outputs in some cases. Graph-LLaDA with $\lambda{=}0.25$ consistently produces fluent sentences that cover all facts in the input graph.

\subsection{Pairwise LLM Win Rates}











\begin{table*}[thb]
\centering
\small
\begin{tabular}{lccccc}
\toprule
\textbf{Model} &
\textbf{Avg. Elo} &
\textbf{Fluency} &
\textbf{Hallucination} &
\textbf{Omission} &
\textbf{Avg. Std} \\
\midrule
GAP (\textsc{WebNLG}) 
& 1226 & 1089 & 1298 & 1292 & 33 \\
GAP (\textsc{LAGRANGE}) 
& 1457 & 1659 & 1302 & 1411 & 31 \\
LLaMA 3.1-8B 
& 1522 & 1460 & 1551 & 1555 & 30 \\
Graph-LLaDA $\lambda{=}1.0$  
& 1547 & 1576 & 1578 & 1487 & 31 \\
LLaMA 3.3-70B 
& 1572 & 1503 & 1588 & \textbf{1624} & 27 \\
Graph-LLaDA $\lambda{=}0.25$ 
& 1581 & 1546 & \textbf{1592} & 1606 & {22} \\
LLaDA-8B 
& \textbf{1595} & \textbf{1667} & \textbf{1592} & 1526 & 27 \\
\bottomrule
\end{tabular}
\caption{Avg. Elo ratings and standard deviations on LAGRANGE across fluency, hallucination, and omission evaluation dimensions, using Gemma~4 and Qwen~3.6 as judges. Best results are bolded.}
\label{tab:elo_score_lagrange}
\end{table*}













\begin{table*}[thb]
\centering
\small
\begin{tabular}{lccccc}
\toprule 
\textbf{Model} & \textbf{Avg. Elo} & \textbf{Fluency} & \textbf{Hallucination} & \textbf{Omission} & \textbf{Avg. Std} \\
\midrule
GAP (\textsc{LAGRANGE}) 
& 1325 & 1396 & 1275 & 1305 & 36 \\
Graph-LLaDA $\lambda{=}1.0$ 
& 1412 & 1428 & 1440 & 1368 & 37 \\
GAP (\textsc{WebNLG}) 
& 1424 & 1413 & 1422 & 1437 & 34 \\
LLaDA-8B $\lambda{=}0.5$ 
& 1489 & 1371 & 1540 & 1555 & 28 \\
LLaMA 3.1-8B 
& 1505 & 1484 & 1515 & 1516 & 31 \\
LLaDA-8B 
& 1513 & 1594 & 1500 & 1443 & 32 \\
LLaDA-8B $\lambda{=}1.0$ 
& 1513 & 1451 & 1535 & 1554 & 28 \\
Graph-LLaDA $\lambda{=}0.25$ 
& 1572 & 1601 & 1553 & 1562 & {27} \\
LLaMA 3.3-70B 
& \textbf{1605} & \textbf{1679} & \textbf{1564} & \textbf{1572} & 27 \\
\bottomrule
\end{tabular}
\caption{Elo ratings and standard deviations on WebNLG across fluency, hallucination, and omission evaluation dimensions, using Qwen~3.6 as judges. GAP are trained on LAGRANGE and evaluated on WebNLG. Best results are bolded.}
\label{tab:elo_score_webnlg}
\end{table*}
\begin{table}[thb]
\centering
\small
\setlength{\tabcolsep}{7pt}

\begin{tabular}{lcc}
\toprule
\textbf{Metric} & \textbf{Gemma4} & \textbf{Qwen3.6} \\
\midrule

\multicolumn{3}{l}{\textit{Fluency}} \\
Pos-1 Win (\%)                 & 52.47 & 51.80 \\
Pos-2 Win (\%)                 & 39.08 & 31.49 \\
Model Consistency (\%)         & 73.47 & 59.58 \\
Decisive Pair Preference (\%)  & 18.88 & 24.15 \\
\midrule

\multicolumn{3}{l}{\textit{Faithfulness}} \\
Pos-1 Win (\%)                 & 29.60 & 26.51 \\
Pos-2 Win (\%)                 & 25.86 & 23.61 \\
Model Consistency (\%)         & 42.00 & 44.06 \\
Decisive Pair Preference (\%)  & 8.55  & 4.47 \\
\midrule

\multicolumn{3}{l}{\textit{Coverage}} \\
Pos-1 Win (\%)                 & 42.17 & 36.73 \\
Pos-2 Win (\%)                 & 22.94 & 26.70 \\
Model Consistency (\%)         & 42.86 & 51.48 \\
Decisive Pair Preference (\%)  & 22.98 & 11.60 \\
\bottomrule
\end{tabular}

\caption{Positional bias analysis across evaluation dimensions. Position-1 and Position-2 denote display order rather than fixed model identities.}
\label{tab:elo}
\end{table}

\begin{figure*}[thb]
    \centering
   \includegraphics[width=1.0\linewidth]{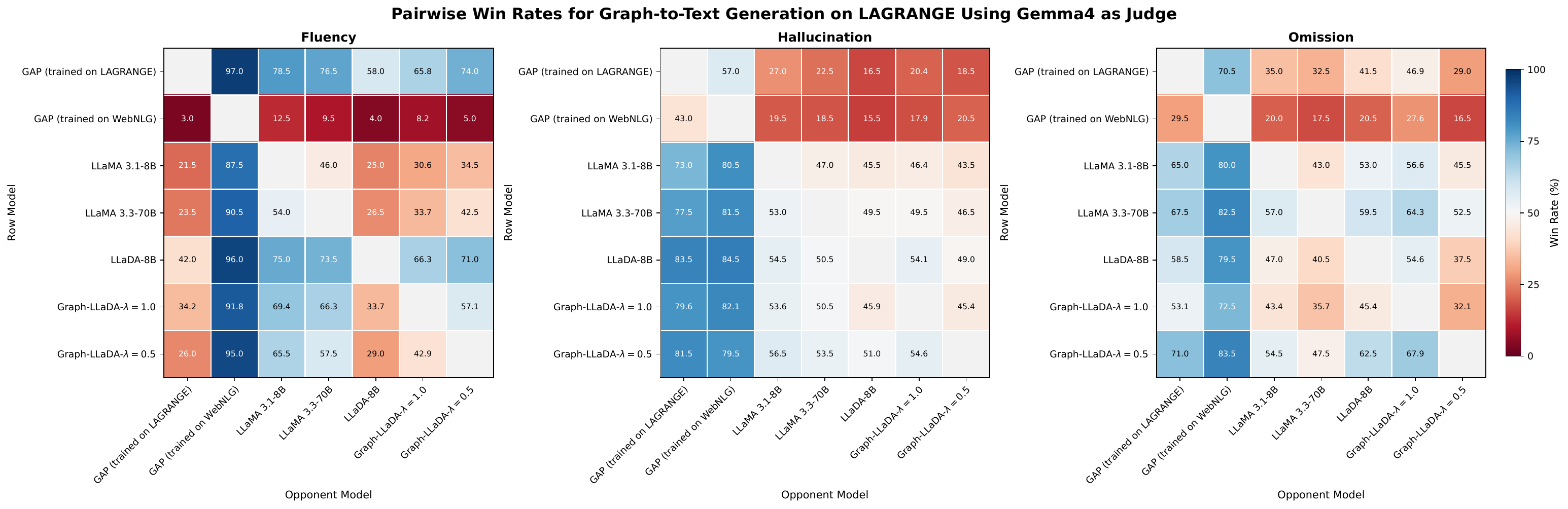}
  \caption{Pairwise Win Rates for Graph-to-Text Generation on LAGRANGE Using Gemma~4-31B as Judge.}
  \label{fig:elo_gemma4_lagrange}
\end{figure*}

\begin{figure*}[thb]
    \centering
   \includegraphics[width=1.0\linewidth]{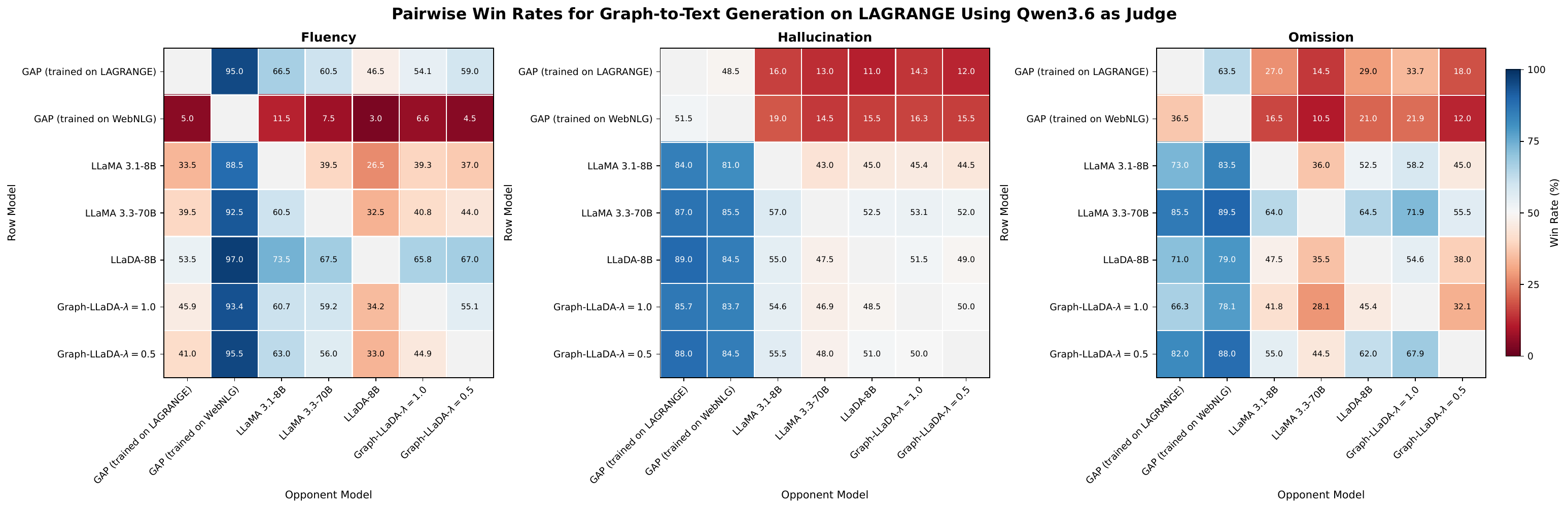}
  \caption{Pairwise Win Rates for Graph-to-Text Generation on LAGRANGE Using Qwen~3.6-35B as Judge.}
  \label{fig:elo_qwen_lagrange}
\end{figure*}

\begin{figure*}[thb]
    \centering
   \includegraphics[width=1.0\linewidth]{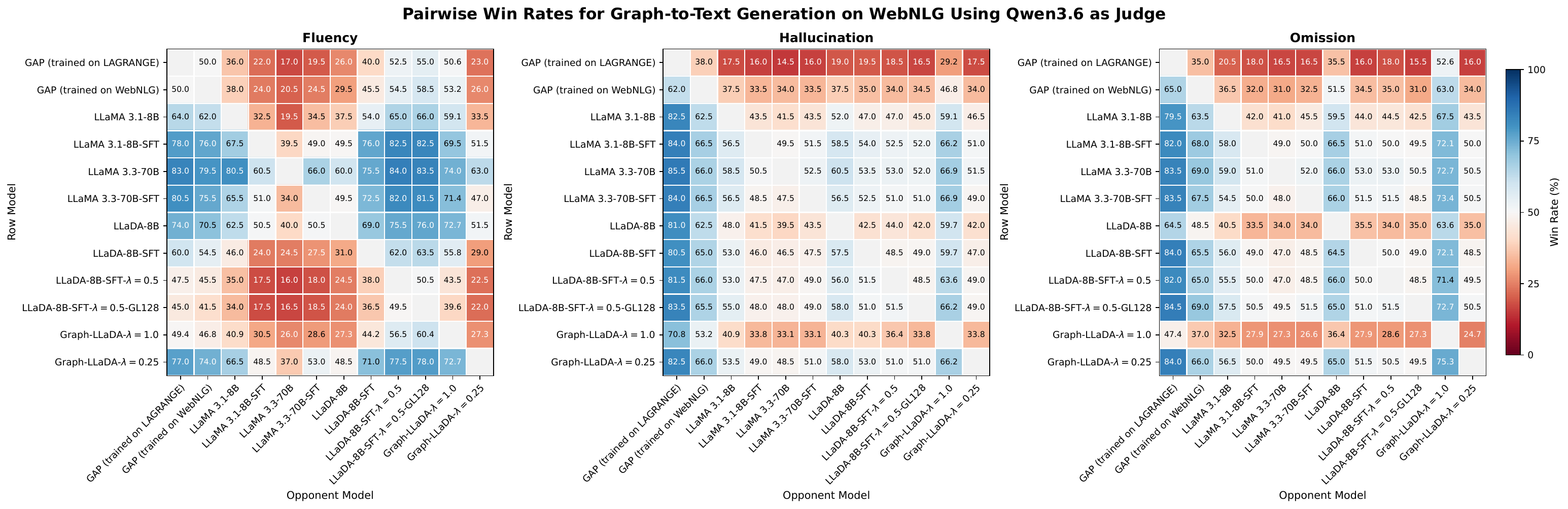}
  \caption{Pairwise Win Rates for Graph-to-Text Generation on WebNLG Using Qwen~3.6-35B as Judge.}
  \label{fig:elo_qwen_webnlg}
\end{figure*}

As shown in Figures~\ref{fig:elo_gemma4_lagrange}, \ref{fig:elo_qwen_lagrange}, and~\ref{fig:elo_qwen_webnlg}, GAP achieves competitive fluency under in-distribution conditions but is consistently outperformed by all other models on faithfulness and coverage---consistent with the automatic metric results discussed in Section~\ref{sec:exp-traj-analysis}. LLaDA similarly scores highly on fluency while tending to omit graph content, as illustrated in Figure~\ref{fig:example}.

Table~\ref{tab:elo} reveals that both Gemma~4 and Qwen~3.6 exhibit substantial positional bias, consistently favoring Position-1 responses across all evaluation dimensions, with the effect most pronounced for fluency. This bias persists after response-order swapping and is weakest for faithfulness, suggesting that more subjective dimensions are disproportionately susceptible to presentation-order effects. We mitigate this by retaining a preference only when the judge's decision is consistent across both orderings, treating inconsistent judgments as ties (Section~\ref{sec:methodology}).

\begin{table}[thb]
\centering
\small
\begin{tabular}{lccc}
\toprule
\textbf{Evaluator} & \textbf{Fluency} & \textbf{Omissions} & \textbf{Hallucinations} \\
\midrule
Human-1 & 90\% & 92\% & 98\% \\
Human-2 & 94\% & 94\% & 96\% \\
Sonnet 4.6   & 100\% & 94\% & 94\% \\
GPT-5.5   & 100\% & 100\% & 96\% \\
\bottomrule
\end{tabular}
\caption{Percentage of positive quality scores for Graph-LLaDA-generated sentences across human annotators, Claude Sonnet~4.6 and GPT~5.5 over the dimensions of fluency, omissions, and hallucinations.}
\label{tab:human-analysis}
\end{table}

\begin{table}[thb]
\centering
\small

\begin{tabular}{lcc}
\toprule
\textbf{Dimension} & \textbf{Agreement} & \textbf{Cohen's Kappa} \\
\midrule

\multicolumn{3}{l}{\textbf{Fluency}} \\
Human1 vs Sonnet 4.6   & 0.900 & 0.000 \\
Human1 vs GPT~5.5  & 0.900 & 0.000 \\
Sonnet 4.6 vs GPT~5.5  & 1.000 & N/A \\
Human2 vs Sonnet 4.6   & 0.940 & 0.000 \\
Human2 vs GPT~5.5  & 0.940 & 0.000 \\
Sonnet 4.6 vs GPT~5.5  & 1.000 & N/A \\
\midrule

\multicolumn{3}{l}{\textbf{Omissions}} \\
Human1 vs Sonnet 4.6   & 0.960 & 0.728 \\
Human1 vs GPT~5.5  & 0.920 & 0.000 \\
Sonnet 4.6 vs GPT~5.5  & 0.920 & 0.000 \\
Human2 vs Sonnet 4.6   & 0.940 & 0.370 \\
Human2 vs GPT~5.5  & 0.940 & 0.000 \\
Sonnet 4.6 vs GPT~5.5  & 0.960 & 0.000 \\
\midrule

\multicolumn{3}{l}{\textbf{Hallucinations}} \\
Human1 vs Sonnet 4.6   & 0.920 & -0.031 \\
Human1 vs GPT~5.5  & 0.940 & -0.027 \\
Sonnet 4.6 vs GPT~5.5  & 0.980 & 0.790 \\
Human2 vs Sonnet 4.6   & 0.900 & -0.050 \\
Human2 vs GPT~5.5  & 0.920 & -0.042 \\
Sonnet 4.6 vs GPT~5.5  & 0.980 & 0.790 \\
\bottomrule

\end{tabular}

\caption{Inter-rater agreement and Cohen's Kappa scores for fluency, omissions, and hallucination evaluations of Graph-LLaDA-generated sentences across human annotators, Sonnet 4.6 Sonnet~4.6, and GPT~5.5.}
\label{tab:agreement_kappa}

\end{table}


\section{Extended Discussion}
\label{app:discussion}

\subsection{Alternatives to $\lambda$-Scaled Structural Decoding}
\label{app:alternatives}

Before settling on $\lambda$-scaled structural decoding, we considered three other inference-time interventions for the early-structural-commitment problem. We summarize them here for completeness; each is straightforward to implement, but none addressed the root cause as cleanly as $\lambda$-scaling.

\paragraph{Confidence-threshold remasking.}
A natural baseline---and the strategy underlying Mask-Predict~\cite{ghazvininejad-etal-2019-mask}---is to refuse to commit any token whose confidence falls below a fixed threshold, regardless of whether it is in the top-$k$ for the step. This does not address our failure mode: by construction, the structural tokens we want to delay are \emph{exactly the tokens whose confidence is highest} early in the trajectory (Table~\ref{tab:llada_metrics_transposed}). A confidence threshold thus has no effect on premature period/EOS placement unless it is set so high that legitimate content tokens are also rejected, in which case the trajectory grinds to a halt. $\lambda$-scaling instead directly downweights structural-token confidence so they no longer dominate the top-$k$ ranking.

\paragraph{Position-conditional commitment.}
A second alternative is to forbid commitment of structural tokens until some fraction of positions have been unmasked---e.g., ``never commit a period while more than $32$ positions remain masked.'' This works in principle but has the same flavor of arbitrariness one usually wants to avoid: the threshold is in absolute positions, so it does not generalize across generation lengths (a threshold tuned for $\mathit{gl}=128$ is wrong for $\mathit{gl}=32$ or $\mathit{gl}=256$), and the boundary between ``allowed'' and ``forbidden'' is a step function rather than a soft penalty. $\lambda$-scaling replaces this with a single dimensionless scalar that scales naturally across generation lengths.

\paragraph{Entropy-aware commitment.}
A third alternative is to commit a structural token contingent on low entropy of the model's distribution over non-structural alternatives at that position---intuitively, `placing a period only when the model is confident there is nothing else to say. We initially found this attractive because it appears principled, but on reflection, it is operationally close to $\lambda$-scaling: in both cases, structural tokens require a higher implicit bar to be committed, and the bar is satisfied roughly when the rest of the sequence has been resolved. The entropy formulation adds a per-position entropy computation and an extra hyperparameter (the entropy threshold) without changing the qualitative effect. We therefore prefer the simpler $\lambda$ formulation.

Together these alternatives clarify the design space and motivate $\lambda$-scaling as the minimal intervention that targets the root cause---inflated confidence on a small set of structural tokens---without depending on absolute positions or auxiliary entropy estimates. We do not claim $\lambda$-scaling is optimal (cf.\ the conclusion), only that it is the simplest fix in this family that we found to work reliably.

\subsection{Future Work}
\label{app:future-work}

We view $\lambda$-scaled structural decoding as a diagnosis-driven workaround rather than a final solution to premature structural commitment. Its optimal value depends on training quality, it requires a per-checkpoint sweep, and it modifies inference rather than fixing the underlying calibration of the SFT-tuned model.

More principled alternatives include training-time fixes such as flexible-length MDLMs~\cite{kim2025any}, calibration losses targeting structural tokens specifically, or learned remasking schedules that adapt to the unmasking trajectory. Investigating these, and extending the SFT failure-mode analysis to other MDLMs (Dream-7B, LLaDA-MoE, LLaDA~2)---we have only confirmed the base-model entity-first pattern on Dream---are the most natural follow-ups to this work.

\end{document}